%% file: main.tex
\documentclass[10pt,twocolumn,letterpaper]{article}

\usepackage[pagenumbers]{cvpr} 

\input{preamble}
\usepackage{xcolor}         
\usepackage{graphicx}
\usepackage{multirow}
\usepackage{bbding}
\usepackage{array}
\usepackage{colortbl}
\usepackage{enumitem}
\usepackage{amsmath}
\usepackage{tcolorbox}
\tcbuselibrary{breakable}
\usepackage{mathrsfs}
\usepackage{booktabs}
\usepackage{tabularx}
\usepackage{array}
\newcolumntype{Y}{>{\raggedright\arraybackslash}X}
\definecolor{mygreen}{HTML}{619129}
\definecolor{myblue}{HTML}{498CC5}

\definecolor{cvprblue}{rgb}{0.21,0.49,0.74}
\usepackage[pagebackref,breaklinks,colorlinks,allcolors=cvprblue]{hyperref}

\def\paperID{4499} %
\def\confName{CVPR}
\def\confYear{2026}

\title{Global Prior Meets Local Consistency: Dual-Memory Augmented Vision-Language-Action Model for Efficient Robotic Manipulation}

\author{
Zaijing Li$^{1\,2}$,
Bing Hu$^{1}$,
Rui Shao$^{1\,3}$\footnotemark[1], Gongwei Chen$^{1}$, Dongmei Jiang$^{2}$\footnotemark[1], \\Pengwei Xie$^{4}$, Jianye HAO$^{4}$, Liqiang Nie$^{1}$
\\ 
    $^1$Harbin Institute of Technology, Shenzhen  \quad $^2$PengCheng Laboratory \\
      $^3$Shenzhen Loop Area Institute \quad $^4$Huawei Noah's Ark Lab\\
       \href{https://cybertronagent.github.io/OptimusVLA.github.io/}{https://cybertronagent.github.io/OptimusVLA.github.io/}
  }

\begin{document}
\maketitle
\renewcommand{\thefootnote}{\fnsymbol{footnote}} 
\input{sec/0_abstract} 
\renewcommand{\thefootnote}{\arabic{footnote}}
\input{sec/1_intro}

\input{sec/2_related}
\input{sec/3_method}

\input{sec/4_exp}

\input{sec/5_conclusion}

{
    \small
    \bibliographystyle{ieeenat_fullname}
    \bibliography{main}
}

\clearpage
\input{Supplementary/X_suppl}
\newpage


\end{document}

%% file: sec/0_abstract.tex
\begin{abstract}
Hierarchical Vision–Language–Action (VLA) models have rapidly become a dominant paradigm for robotic manipulation. It typically comprising a Vision–Language backbone for perception and understanding, together with a generative policy for action generation. However, its performance is increasingly bottlenecked by the action generation proceess. (i) Low inference efficiency. A pronounced distributional gap between isotropic noise priors and target action distributions, which increases denoising steps and the incidence of infeasible samples. (ii) Poor robustness. Existing policies condition solely on the current observation, neglecting the constraint of history sequence and thus lacking awareness of task progress and temporal consistency. To address these issues, we introduce \textbf{OptimusVLA}, a dual-memory VLA framework with \textbf{G}lobal \textbf{P}rior \textbf{M}emory (\textbf{GPM}) and \textbf{L}ocal \textbf{C}onsistency \textbf{M}emory (\textbf{LCM}). GPM replaces Gaussian noise with task-level priors retrieved from semantically similar trajectories, thereby shortening the generative path and reducing the number of function evaluations (NFE). LCM dynamically models executed action sequence to infer task progress and injects a learned consistency constraint that enforces temporal coherence and smoothness of trajectory. Across three simulation benchmarks, OptimusVLA consistently outperforms strong baselines: it achieves 98.6\% average success rate on LIBERO, improves over $\pi_{0}$ by 13.5\% on CALVIN, and attains 38\% average success rate on RoboTwin 2.0 Hard. In Real-World evaluation, OptimusVLA ranks best on Generalization and Long-horizon suites, surpassing $\pi_{0}$ by 42.9\% and 52.4\%, respectively, while delivering 2.9× inference speedup. 
\end{abstract}


%% file: sec/1_intro.tex
\section{Introduction}
\label{intro}
\input{figures/fig-1}

Vision–Language–Action (VLA) models \cite{kim2024openvla,li2025cogvla,zhang2025dreamvla,wang2025vla-adapter,song2025pdvla,qu2025spatialvla,li2025semanticvla,black2024pi0,intelligence2025pi05,zhu2026H-GAR} are rapidly emerging as a dominant paradigm for robotic manipulation \cite{jiang2022vima,li2023roboflamingo,song2026energyaction}. As shown in Fig. \ref{fig:fig1} Top, it integrates observation perception, natural-language understanding, and action generation within an end-to-end framework. Leveraging advances in vision–language models \cite{liu2024visual,chen2024lion} and generative policies \cite{chi2025diffusionpolicy,zhang2025flowpolicy,li2025star}, VLA models have demonstrated strong performance across a variety of benchmarks \cite{liu2023libero,mees2022calvin,chen2025robotwin2}. However, the action generation process has become a dominant bottleneck in terms of both efficiency and robustness.

Limitation \uppercase\expandafter{\romannumeral 1}: Inefficient action generation due to a large prior–target gap. As shown in Fig. \ref{fig:fig1} Middle, mainstream VLA models map a prior noise distribution (typically Gaussian) to the target action distribution via diffusion denoising or flow matching \cite{intelligence2025pi05,pertsch2025pi0fast}. However, the cross-domain transformation from isotropic noise to structured actions is large, necessitating multi-steps denoising to reach high(-quality actions \cite{liu2022flow}. Moreover, the stochastic starting point frequently initializes the generative process in regions that  kinematically invalid actions \cite{ma2024hierarchical,Ortenzi2018kinematics}. While a naive strategy might use \emph{action-prior} as starting point, this approach severely curtails diversity, collapsing the learned mapping to a restrictive \emph{similar-to-target} function and failing to address how such priors are obtained.

Limitation \uppercase\expandafter{\romannumeral 2}: Poor robustness to temporal dependence. VLA models \cite{kim2024openvla,intelligence2025pi05} rely solely on the current observation often fails to distinguish between distinct task phases that yield similar visual inputs (e.g., distinguishing an unopened drawer from one that has just been closed). Moreover, lacking temporal dependency weakens consistency with the executed trajectory, leading to jittery control. While a straightforward solution \cite{liu2025robovlms,fan2025interleave-vla} is to concatenate long-sequence historical observation to the input, it substantially increases inference overhead and memory usage. Moreover, it misaligned the pre-training distribution of VLA models \cite{shi2025memoryvla}.

To address these issues, we propose \textbf{OptimusVLA}, a VLA framework jointly driven by \textbf{G}lobal \textbf{P}rior \textbf{M}emory (\textbf{GPM}) and \textbf{L}ocal \textbf{C}onsistency \textbf{M}emory (\textbf{LCM}). \textbf{GPM} replaces isotropic noise with task-level priors as the generative start, narrowing the \emph{prior–target} distributional gap and reducing required number of function evaluations (NFE). \textbf{LCM} dynamically models recent action sequences to infer task progress and injects a consistency constraint that smooths controls. Together, GPM and LCM improve efficiency and robustness of OptimusVLA through global prior alignment and local consistency constraints.

The core innovation of GPM is to treat prior initialization as a memory-driven retrieval problem rather than a fixed noise design.  It consists of three components: a Prior Head that integrates current multimodal information; a Memory Bank that stores and retrieves task-level priors; and a Prior-Aware Sampler for prior action distribution sampling together with adaptive noise scale and NFE. By initializing the generative flow in the neighborhood of similar tasks, GPM drastically narrows the prior-target gap. This not only reduce the required NFE but, critically, anchors the process in feasible action space. Adaptive noise injection retains the model's exploratory power, preventing deterministic collapse and preserving generalization to novel tasks.

Concurrently, the LCM endows the VLA model with temporal perception without the burden of long context modeling. It employs two lightweight structures: a Consistency Layer that ingests recent action chunks to model local consistency, and a Dynamic-Awareness module that dynamicaly models historical sequence to infer task progress. The LCM learns a consistency constraint that injected into the policy input, without requiring any modification to the VLA pre-training paradigm. It enforces consistency with historical trajectory, and provides crucial progress awareness with negligible computational overhead.

We conduct comprehensive evaluation across three simulators and the real world. On LIBERO \cite{liu2023libero}, OptimusVLA attains an average success rate of 98.6\%, surpassing current SOTA baselines (e.g., $\pi_{0.5}$ \cite{intelligence2025pi05} and MemoryVLA \cite{shi2025memoryvla}). On CALVIN \cite{mees2022calvin}, it improves over $\pi_{0}$ by 13.5\%, and it achieves 38\% average success rate on RoboTwin 2.0 \cite{chen2025robotwin2} \textit{Hard} setting. In real-world evaluations, OptimusVLA ranks best on both the \textit{Generalization} and \textit{Long-horizon} suites, outperforming $\pi_{0}$ by 42.9\% and 52.4\%, respectively, while delivering 2.9$\times$ inference speedup. These results indicate that memory-driven prior initialization and temporal constraint can boost robustness of VLA models without sacrificing efficiency. Our contributions are as follows:

\begin{itemize}
    \item A novel \textbf{G}lobal \textbf{P}rior \textbf{M}emory that narrows the prior-target action distribution gap. By replacing isotropic noise with retrieved task-level prior, GPM significantly reduces NFE and lowers the risk of infeasible sampling.
    \item A lightweight \textbf{L}ocal \textbf{C}onsistency \textbf{M}emory that endows VLAs with temporal awareness. LCM dynamically models recent actions to inject a consistency constraint, providing progress-awareness without appreciable computational overhead.
    \item We propose a dual-memory VLA framework, OptimusVLA, which driven by GPM and LCM. Extensive experiments results across 3 simulators and real-world show that OptimusVLA achieves higher performance with substantially inference speedup.
\end{itemize}

%% file: figures/fig-1.tex
\begin{figure}[htbp]
    \centering
    \includegraphics[width=0.5\textwidth]{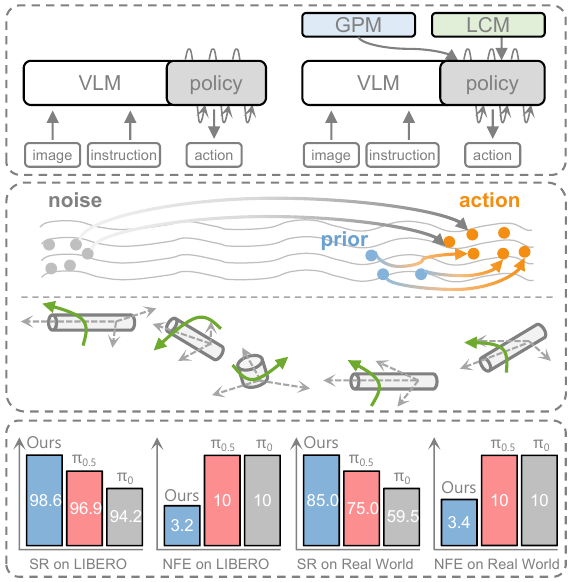}
    \caption{\textbf{Top}: Comparison between the standard VLA architecture (left) and our proposed OptimusVLA (right). (ii) Poor robustness to temporal dependence. \textbf{Middle}: Illustration of how GPM (\textcolor{myblue}{blue}) and LCM (\textcolor{mygreen}{green}) address two key limitations of existing VLA models: (i) Low inference efficiency due to a large prior–target gap. (ii) Poor robustness to temporal dependence. \textbf{Bottom}: Efficiency and performance comparison.}
    \label{fig:fig1}
\end{figure}

%% file: sec/2_related.tex
\section{Preliminaries and Related Work}
\input{figures/fig-2}
Contemporary Vision–Language–Action (VLA) models typically divided into two categories: single-stream architectures \cite{kim2024openvla,kim2025openvla-oft,zitkovich2023rt2,qu2025spatialvla,liu2025hybridvla} and hierarchical architectures \cite{li2024cogact,black2024pi0,pertsch2025pi0fast,intelligence2025pi05,deng2025graspvla}. Single-stream architectures \cite{zhou2025hiconagent,li2026optimus3,xiang2026tina} employ a vision–language backbone and autoregressively generate discretized action tokens \cite{huang2023leo,jiang2022vima}. To meet the high-frequency control demands of real-world scenarios, various efficiency techniques, such as parallel decoding \cite{kim2025openvla-oft,song2025pdvla}, parameter quantization \cite{wang2025bitvla}, and token compression \cite{li2025cogvla,li2023roboflamingo}, have been adopted to increase inference speed. On the other hand, hierarchical approaches \cite{li2024optimus,li2025optimus2,zhang2025falcon} contain a vision–language backbone for high-level, low-frequency planning, a generative policy \cite{chi2025diffusionpolicy,zhang2025flowpolicy} for low-level, high-frequency action generation \cite{black2024pi0,intelligence2025pi05,wen2025tinyvla}. This decomposition substantially accelerates inference compared with purely autoregressive pipelines. Nevertheless, even with flow matching \cite{lipman2022flow,liu2022flow}, the policy typically requires multiple NFEs to reach high quality actions, which ultimately caps the achievable speedup.

Conditional Flow Matching (CFM) trains a time-conditioned velocity field $v_\theta(x,t)$ that transports the noise $x_0\sim \mathcal{P}_{0}$ to the target action $x_1\sim \mathcal{P}_{1}$. A common and stable approach is to define an Optimal Transport (OT) path as a straight line:
\begin{equation}
x_t = (1-t)x_0 + t x_1.
\label{eq:ot_path}
\end{equation}
The target velocity $u_t$ along this path is constant, given by $u_t(x_t \mid x_0, x_1) = x_1 - x_0$. The learning objective is to train $v_\theta$ to match this target velocity field:
\begin{equation}
\min_{\theta}\;\; 
\mathbb{E}_{\substack{t\sim\mathcal{U}[0,1],x\sim p_{t}(x)}}
\Big\|\, v_\theta\!\big(t,x) - u_t(x) \,\Big\|_2^2,
\qquad 
\label{eq:cfm_obj}
\end{equation}
When $\mathcal{P}_{0}$ is isotropic Gaussian distribution and $\mathcal{P}_{1}$ is a structured action distribution, the source-target gap can be large, which in practice yields more number of function evaluations (NFE) to reach high-quality actions. To narrow the gap, we introduce Global Prior Memory, which replaces the Gaussian source by a task-level prior constructed from semantically similar action trajectories.

On the other hand, mainstream work \cite{kim2024openvla,black2024pi0,intelligence2025pi05} formulates action generation in VLA as a Markov Decision Process, conditioning the model solely on the current observation while ignoring the historical sequence. This Markovian assumption hampers phase awareness: the agent cannot discern task progress from visually similar states (e.g., whether the drawer has not yet been opened or has already been closed), leading to suboptimal or inconsistent behaviors \cite{shi2025memoryvla}. To provide temporal context, some work \cite{liu2025robovlms,fan2025interleave-vla} adapt the VLM paradigm to sequence modeling, explicitly encoding long observation streams. While effective, these approaches substantially increase latency and memory usage. Other methods simply concatenate past states \cite{zheng2024tracevla} or actions \cite{bu2025univla} to the input, which lacks a global representation of long-range temporal context \cite{zhang2023attribute,zhang2025spatial,li2022emocaps,zhang2024multi}. Moreover, it misaligned the VLA model's single-frame pre-training distribution \cite{kim2024openvla}. A different line of work employs working memory \cite{li2025optimus2,shi2025memoryvla} to dynamically model observation history. However, these designs typically rely on the VLM backbone to represent the current multimodal information at every step; consequently, each memory update incurs a full VLM forward pass, creating a throughput bottleneck for fast control loops. In this paper, we propose Local Consistency Memory (LCM), a lightweight working memory that dynamically encodes the recent sequence with minimal inference overhead. LCM equips VLA policies with explicit progress awareness and temporal consistency without repeatedly invoking heavy VLM computations, thereby preserving real-time efficiency.

%% file: figures/fig-2.tex
\begin{figure*}[htbp]
    \centering
    \includegraphics[width=1.0\textwidth]{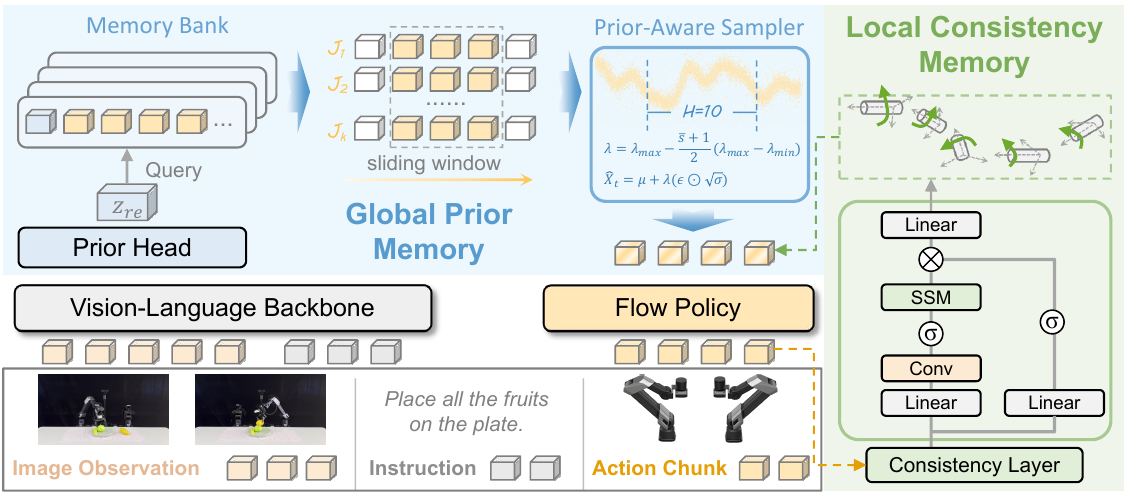}
    \caption{Overview of OptimusVLA framework. Given a task and the current observation, the Vision–Language backbone first encodes the inputs into a multimodal representation. GPM then retrieves a task-level prior based on this representation, while LBM dynamically encodes the historical action sequence to produce a consistency constraint. Finally, the flow policy denoises the initialization with an adaptive NFEs schedule to generate the action chunk.}
    \vspace{-10pt}
    \label{fig:fig2}
\end{figure*}

%% file: sec/3_method.tex
\section{OptimusVLA}
In this section, we first give an overview of our proposed OptimusVLA. Next, we introduce the details of \textbf{G}lobal \textbf{P}rior \textbf{M}emory (\textbf{GPM}) in Sec. \ref{gpm}. Subsequently, we elaborate on how to implement the proposed \textbf{L}ocal \textbf{C}onsistency \textbf{M}emory (\textbf{LCM}) (Sec. \ref{LCM}). Finally, the training strategy of the OptimusVLA is explained in Sec \ref{training}.

\subsection{Overview of OptimusVLA}
\label{overview}
 As shown in Figure \ref{fig:fig2}, OptimusVLA comprises Vision–Language backbone \cite{liu2024visual,chen2024lion}, flow policy \cite{zhang2025flowpolicy}, Global Prior Memory (GPM), and Local Consistency Memory (LCM). The Vision–Language backbone and flow policy are the standard components of a hierarchical VLA architecture. GPM explicitly tackles the distributional gap between the standard noise prior and the structured action distribution, while LCM injects progress-aware temporal constraints, jointly enhancing efficiency and robustness of action generation.

Formally, given a natural-language instruction 
$\ell$ and the current observation $O_t$, the multimodal representation $E_{emb}$ is embedded by Vision-Language backbone \texttt{VLM}: 
\begin{equation}
E_{emb} \gets \texttt{VLM} ( O_t, \ell  )
\end{equation}
where $O_{t}$ typically consists of the proprioceptive state $q_{t}$, and multi-view images 
$I^{1}_{t}, I^{2}_{t}, \dots, I^{n}_{t}$.

Subsequently, the multimodal representation $E_{emb}$ is projected into a retrieval token $z_{re}$, which queries the memory bank in GPM to obtain a task-level prior action distribution $\mathcal{P}_{re}$:
\begin{equation}
\mathcal{P}_{re} \gets \texttt{GPM} ( z_{re}  )
\end{equation}

At time $t$, the GPM samples an action chunk $\mathbf{\hat{X}}_{t} \in \mathbb{R}^{H\times A} \sim \mathcal{P}_{re}$ with adaptive noise and NFE, where $H$ is the chunk length and $A$ is the action dimension. Concurrently, the LCM processes the last action chunk $\mathbf{A}_{t-1}$ to yield a consistency bias $\mathbf{B}_{t} \in \mathbb{R}^{H\times A}$, then this bias is injected to form the final policy input $\mathbf{X}_{t}$:
\begin{equation}
\mathbf{B}_{t} \gets \texttt{LCM} ( \mathbf{A}_{t-1} )
\label{eq:lcm_forward}
\end{equation}
\begin{equation}
\mathbf{X}_{t} = \mathbf{\hat{X}}_{t} + \mathbf{B}_{t}
\label{add_consistency}
\end{equation}

Finally, the flow policy $p_{\theta}$ transforms $\mathbf{X}_{t}$ into action chunk $a_{t+1:t+H}$ based on adaptive NFE $N$:
\begin{equation}
a_{t+1:t+H} \gets p_{\theta} ( \mathbf{X}_{t}, N )
\end{equation}

\subsection{Global Prior Memory}
\label{gpm}
Standard flow matching transports isotropic Gaussian noise to the action space \cite{lipman2022flow,liu2022flow}. Due to the large discrepancy between these distributions, this process requires many NFEs and risks collapsing to infeasible actions. In robotics, however, similar tasks share related action distributions (e.g., \texttt{pick\_a\_cup} and \texttt{pick\_a\_plate}). Motivated by this, our \textbf{G}lobal \textbf{P}rior \textbf{M}emory (\textbf{GPM}) stores, retrieves, and composes trajectories into task-level priors. This fundamentally relocates the generative starting point from $\mathcal{N}(0, I)$ to the neighborhood of the target manifold, thereby narrowing the prior-target gap.

GPM is a long-term memory module composed of a Prior Head, a Memory Bank, and a Prior-Aware Sampler. 

\noindent\textbf{Prior Head}. A lightweight MLP projects the multimodal representation $E_{emb}$ into a retrieval token $z_{re}$:
\begin{equation}
z_{re} = PriorHead (E_{emb} )
\end{equation}

\noindent\textbf{Memory Bank}. $z_{re}$ queries the Memory Bank, which stores $M$ key-value pairs $\{z_{m}, J_{m}\}_{m=1}^{M}$ of task embeddings and their corresponding full trajectories. We compute cosine similarities and retrieve the $k$ nearest trajectories $\{J_{i}\}_{i=1}^{k}$ and their scores $s_i$:
\begin{equation}
\{J_i, s_i\}_{i=1}^k \;\gets\; \mathrm{MemoryBank}\!\left(z_{re}\right)
\end{equation}
Then we compute a normalized global similarity $\bar s$ and softmax weights $\alpha_i$:
\begin{equation}
\alpha_i = \mathrm{softmax}(s_i / \tau_s), \quad
\bar s=\sum_{i=1}^{k}\alpha_i s_i,
\end{equation}
where $\tau_s$ is a temperature. For each retrieved trajectory $J_i$, we extract action blocks $C_i \in \mathbb{R}^{H\times A}$ based on a sliding window. The task-level prior $\mathcal{P}_{re} = \mathcal{N}(\mu, \mathrm{diag}(\mathrm{Var}))$ is then formed by a weighted average:
\begin{equation}
\mu=\sum_{i=1}^{k}\alpha_i C_i,\qquad
\mathrm{Var}=\sum_{i=1}^{k}\alpha_i\, (C_i-\mu)^{\odot 2},
\label{eq:ltm-stats}
\end{equation}
where $\odot 2$ denotes element-wise power.

\noindent\textbf{Prior-Aware Sampler}. Subsequently, Prior-Aware Sampler samples the initialization with similarity-adaptive noise $\lambda$, and adaptive NFE $N$:
\begin{equation}
\lambda=\lambda_{\max}-\tfrac{\bar s+1}{2}\,(\lambda_{\max}-\lambda_{\min}),
\end{equation}
\begin{equation}
N=N_{\min}+\big(1-\tfrac{\bar s+1}{2} \big)\,(N_{\max}-N_{\min}),
\label{eq:ltm-schedules}
\end{equation}
\begin{equation}
\mathbf{\hat{X}}_{t}=\mu+\lambda\,\big(\epsilon\odot\sqrt{\mathrm{Var}}\big), \qquad
\epsilon\sim\mathcal{N}(0,I),
\end{equation}
where $\lambda_{\min}$, $\lambda_{\max}$, $N_{\min}$, $N_{\max}$ are hyper-parameters.

GPM draws the prior distribution into the vicinity of the target distribution, thereby drastically curtailing the required NFEs and improving the inference efficiency of the flow policy. Moreover, by initializing from a reliable prior, it reduces the likelihood that sampled actions fall into infeasible regions.

\input{table/tb_main_libero}

\subsection{Local Consistency Memory}
\label{LCM}
Standard VLAs, adhering to the Markovian assumption, lack explicit progress awareness \cite{shi2025memoryvla} and suffer from jittery control. Our \textbf{L}ocal \textbf{C}onsistency \textbf{M}emory (LCM) addresses this by dynamically modeling the action history to infer task progress and enforce temporal consistency, all with minimal computational overhead. LCM serves as a working memory, comprising a Consistency Layer and a Dynamic Awareness Module. 

\noindent\textbf{Consistency Layer}. At time step $t$, the Consistency Layer incorporates the previous action chunk $\mathbf{A}_{t-1}=[\mathbf{a}_{t-H+1},\ldots,\mathbf{a}_t]\in\mathbb{R}^{H\times A}$ and employs a self-attention mechanism \cite{attention2017} to capture inter-action dependencies and constraints within the chunk:
\begin{equation}
\mathbf{\hat{B}}_{t-1} \gets  ConsistencyLayer(\mathbf{A}_{t-1})
\end{equation}

\noindent\textbf{Dynamic Awareness Module}. This module is designed to capture inter-chunk temporal dynamics. It takes the representation $\mathbf{\hat{B}}_{t-1}$ and updates its internal state to predict the consistency bias $\mathbf{B}_{t}$ for the next time step (as in Eq. \ref{eq:lcm_forward}). We implement this module using a Mamba-based structure \cite{gu2024mamba}, which efficiently models long-range dependencies with linear complexity:
\begin{equation}
\mathbf{B}_{t} \gets DynamicAwareness(\mathbf{\hat{B}}_{t-1})
\end{equation}

LCM focuses on the temporal dependencies of the current trajectory, converting consistency in the action flow into an additive constraint, thereby yielding smoother sampled trajectories.

\input{table/tb_main_calvin}
\subsection{Training Details}
\label{training}
The training pipeline consists of three stages. First, we pretrain a hierarchical VLA model based on the architecture and training protocol of $\pi_{0.5}$ \cite{intelligence2025pi05}. Next, we train the Prior Head to learn task-discriminative representations from the batch via InfoNCE objective:
\begin{equation}
\mathcal{L}_{\mathrm{GPM}} = -\mathbb{E}_q \left[ \log \frac{\exp(\mathrm{sim}(z_{re}, z^+)/\tau_c)}{\sum_{j \in \mathcal{N}(q)} \exp(\mathrm{sim}(z_{re}, z_j)/\tau_c)} \right]
\end{equation}
where $\mathrm{sim}(\cdot)$ denotes cosine similarity, $\tau_c$ is a temperature parameter, and $\mathcal{N}(q)$ denotes the set of in-batch negative samples. Finally, we freeze GPM and train LCM to predict the residual required to bridge the gap between the global prior mean $\mu_t$ and the ground-truth action chunk $\mathbf{A}^\star_t$: 
\begin{equation}
\mathbf{B}^\star_t = \mathbf{A}^\star_t - \mu_t
\label{eq:lcm_target}
\end{equation}
\begin{equation}
\mathcal{L}_{\mathrm{LCM}} = \mathbb{E}_{(\mathbf{A}_{t-1}, \mathbf{A}^\star_t, \mu_t) \sim \mathcal{D}} \left[ \|\mathbf{B}_t - \mathbf{B}^\star_t\|_2^2 \right]
\label{eq:loss_lcm}
\end{equation}
where $\mathbf{A}^\star_t$ and $\mu_t$ are the ground-truth action chunk and mean prior at timme $t$. More details are in \textbf{Appendix}.

%% file: table/tb_main_libero.tex

\begin{table}[t]
\centering
\caption{Performance comparison on LIBERO~\cite{liu2023libero}. 
We report the average success rate on each task suite (500 rollouts).}
\label{tb:main_libero}
\small
\setlength{\tabcolsep}{4pt}  
\resizebox{\linewidth}{!}{
\begin{tabular}{l|ccccc}
\toprule[1.2pt]
\multirow{2}{*}{\textbf{Method}} 
  & \multicolumn{5}{c}{\textbf{LIBERO}} \\ 
\cmidrule(lr){2-6}
 & Spatial & Object & Goal & Long & Avg. \\ 
\midrule
DP~\cite{chi2025diffusionpolicy}           & 78.3 & 92.5 & 68.3 & 50.5 & 72.4 \\
Octo~\cite{team2024octo}                  & 78.9 & 85.7 & 84.6 & 51.1 & 75.1 \\
SpatialVLA~\cite{qu2025spatialvla}        & 88.2 & 89.9 & 78.6 & 55.5 & 78.1 \\
$\pi_{0}$-FAST~\cite{pertsch2025pi0fast}  & 96.4 & 96.8 & 88.6 & 60.2 & 85.5 \\
CogACT~\cite{li2024cogact}                & 97.2 & 98.0 & 90.2 & 88.8 & 93.6 \\
MemoryVLA~\cite{shi2025memoryvla}         & 98.4 & 98.4 & 96.4 & 93.4 & 96.7 \\
OpenVLA-OFT~\cite{kim2025openvla-oft}     & 97.6 & 98.4 & 97.9 & \underline{94.5} & \underline{97.1} \\
OpenVLA~\cite{kim2024openvla}             & 84.7 & 88.4 & 79.2 & 53.7 & 76.5 \\
UniVLA~\cite{bu2025univla}                & 95.4 & \underline{98.8} & 93.6 & 94.0 & 95.4 \\
$\pi_{0}$~\cite{black2024pi0}             & 96.8 & \underline{98.8} & 95.8 & 85.2 & 94.2 \\
$\pi_{0.5}$~\cite{intelligence2025pi05}   & \underline{98.8} & 98.2 & \underline{98.0} & 92.4 & 96.9 \\
\midrule
\rowcolor[HTML]{E7EEFE}
OptimusVLA                                & \textbf{99.6} & \textbf{99.8} & \textbf{98.4} & \textbf{96.4} & \textbf{98.6} \\
\bottomrule[1.2pt]
\end{tabular}%
}
\end{table}

%% file: table/tb_main_calvin.tex
\begin{table}[t]
\centering
\caption{Performance comparison on CALVIN \cite{mees2022calvin}. 
We report the success rate of each track and average completion length (Avg.\ Len). $^{\dag}$ represents the result we reproduced.}
\label{tb:main_calvin}
\small

\setlength{\tabcolsep}{4pt}  

\resizebox{\linewidth}{!}{
\begin{tabular}{l|cccccc}
\toprule[1.2pt]
\multirow{2}{*}{\textbf{Method}} 
  & \multicolumn{6}{c}{\textbf{CALVIN} (ABC $\rightarrow$ D)} \\ 
\cmidrule(lr){2-7}
 & 1/5 & 2/5 & 3/5 & 4/5 & 5/5 & Avg.\ Len\\ 
\midrule
RoboVLM \cite{liu2025robovlm}        & \textbf{98.0} & \textbf{93.6} & 85.4 & 77.8 & 70.4 & 4.25\\
ReconVLA \cite{song2025reconvla} & 95.6 & 87.6 & 76.9       & 69.3 & 64.1 & 3.95\\
OpenVLA \cite{kim2024openvla} & 91.3 & 77.8 & 62.0 & 52.1 & 43.5 & 3.27\\
UniVLA \cite{bu2025univla} & 95.5 & 85.8 & 75.4 & 66.9 & 56.5 & 3.80\\
UP-VLA \cite{zhang2025up} & 92.8 & 86.5 & 81.5 & 76.9 & 69.9 & 4.08\\
RoboDual \cite{bu2024towards} & 94.4 & 82.7 & 72.1 & 62.4 & 54.4 & 3.66\\
Seer \cite{tian2024predictive} & 96.3 & 91.6 & 86.1 & 80.3 & 74.0 & 4.28\\
VPP \cite{hu2024video} & 95.7 & 91.2 & \underline{86.3} & \underline{81.0} & 75.0 & \underline{4.29}\\
$\pi_{0}$ \cite{black2024pi0} & 93.8 & 85.0 & 76.7 & 68.1 & 59.9 & 3.92\\
$\pi_{0.5}$$^{\dag}$ \cite{intelligence2025pi05} & 94.4 & 88.4 & 85.3 & 80.1 & \underline{76.1} & 4.26\\
\midrule
\rowcolor[HTML]{E7EEFE}
OptimusVLA & \underline{97.6} & \underline{93.2} & \textbf{88.8} & \textbf{85.7} & \textbf{78.1} & \textbf{4.45}\\
\bottomrule[1.2pt]
\end{tabular}%
} 

\end{table}

%% file: sec/4_exp.tex
\section{Experiments}
We conduct comprehensive experiments to validate OptimusVLA, focusing on three pivotal research questions:
\textbf{Q1}. Does OptimusVLA outperform state-of-the-art VLA models on diverse simulation benchmarks?
\textbf{Q2}. Does OptimusVLA demonstrate superior efficiency and performance in real-world?
\textbf{Q3}. How do GPM and LCM individually contribute to the performance of OptimusVLA?

\subsection{Evaluation on Simulation Benchmarks}
\input{table/tb_main_robotwin}

\noindent\textbf{Simulation Benchmarks}. To evaluate the robustness of OptimusVLA across diverse simulation environments, we conduct experiments on LIBERO \cite{liu2023libero}, CALVIN \cite{mees2022calvin}, and RoboTwin 2.0 \cite{chen2025robotwin2}. LIBERO is organized into four task suites (Spatial, Object, Goal, and Long) with each contains 10 tasks. We report the average success rate in each suite with 500 rollouts. CALVIN consists of 4 distinct environments, and we train on 3 environments and testing on the held-out one (ABC $\rightarrow$ D). We report the success rate for each track, as well as the average length over 5 tasks, using 500 rollouts. On RoboTwin 2.0, we randomly select 16 tasks for evaluation, and test the models with 100 rollouts under the \textit{Hard} (domain-randomized with clutter, lighting, textures, and height variations) settings.

\noindent\textbf{Implementation Details}. OptimusVLA is initialized from the weights of $\pi_{0.5}$  \cite{intelligence2025pi05}, and then augment it with the proposed GPM and LCM modules, resulting in 3.6B parameters in total. We train OptimusVLA on 8$\times$ NVIDIA A800 GPUs with a global batch size of 512 for 30,000 steps. The learning rates on LIBERO, CALVIN, and RoboTwin are set to 5e-5. More details are provided in the \textbf{Appendix}.
in

\input{table/tb_ablation}
\noindent\textbf{Results on LIBERO}. As shown in Table \ref{tb:main_libero}, OptimusVLA achieves an average success rate of 98.6\% on LIBERO, surpassing existing state-of-the-art methods, including $\pi_{0.5}$ \cite{intelligence2025pi05} and OpenVLA-OFT \cite{kim2025openvla-oft}. In particular, most of VLA models struggle on the LIBERO-Long suite due to error accumulation over long sequences. OptimusVLA mitigates this via GPM, which retrieves task-specific priors to anchor the generative process. By initializing the flow closer to the target manifold, GPM not only stabilizes long-horizon generation but also drastically reduces the NFEs (3.2 for OptimusVLA vs. 10.0 for $\pi_{0.5}$)

\input{table/tb_ab_memory}

\noindent\textbf{Results on CALVIN}. As shown in Table \ref{tb:main_calvin}, OptimusVLA attains an average episode length of 4.45 on CALVIN, outperforming $\pi_{0}$ \cite{black2024pi0} by 13.5\% and surpassing $\pi_{0.5}$ \cite{intelligence2025pi05}. Under the ABC $\rightarrow$ D setting, $\pi_{0}$ relies on task-agnostic Gaussian noise, which is brittle to the distribution shifts inherent in unseen environments. In contrast, OptimusVLA leverages GPM to retrieve semantically similar trajectories from training data as initialization. This \textit{semantic anchoring} allows the model to adapt to novel scenes by deforming a viable  action  prior rather than generating from scratch, thereby exhibiting superior zero-shot robustness.

\noindent\textbf{Results on RoboTwin 2.0}. As shown in Table \ref{tb:main_robotwin}, OptimusVLA achieves the average success rate of 38\% on RoboTwin 2.0 \textit{Hard} setting. In particular, it attains a 58\% success rate on the \textit{Stack Bowls Two} task, outperforming RDT \cite{liu2024rdt} by +28\% success rate. Bimanual manipulation demands high temporal and inter-arm consistency, whereas RDT lack explicit mechanisms for enforcing such dual-arm coherence, leading to a lower performance. In contrast, LCM provides the necessary consistency constraints with OptimusVLA, enforcing smooth, coordinated trajectories.

\input{figures/fig-3}
\input{figures/fig-4}
\subsection{Evaluation on Real-World}
\noindent\textbf{Training and Evaluation Setting}. To validate the effectiveness of the proposed OptimusVLA in real-world robotic manipulation, we evaluate \textit{Generalization Tasks} and \textit{Long-horizon Tasks} on GALAXEA R1 Lite robot. It is a 14-DoF bimanual platform, and we use both wrist-mounted and third-person cameras with an image resolution of 
224×224. For the \textit{Generalization Tasks}, we train on 100–150 expert demonstrations per task and evaluate each task with 50 rollouts (25 with varying lighting conditions, and 25 with scene variations). For the \textit{Long-horizon Tasks}, we use 200–300 expert demonstrations per task for training and conduct 25 rollouts per task for evaluation. In each rollout, the objects are randomly initialized at different positions. More details are provided in \textbf{Appendix}.

\noindent\textbf{Results on Real-world}. As shown in Fig. \ref{fig:fig3}, OptimusVLA achieves average success rates of 85.0\% and 64.0\% on the \textit{Generalization Tasks} and \textit{Long-horizon Tasks}, respectively. In particular, on the \textit{Generalization Tasks}, OptimusVLA exhibits strong robustness to variations in lighting and scenes. We attribute this to GPM, which retrieves task-level priors conditioned on multimodal semantics, thereby rendering the policy robust to visual distractions. On \textit{Long-horizon Tasks} that require bimanual manipulation of multiple objects, OptimusVLA outperforms $\pi_{0}$ by 52.4\%. This showcases the superior long-horizon action stability and coherent bimanual coordination of OptimusVLA.

\input{figures/fig-5}
\subsection{Ablation Study}
To elucidate the benefits brought by the proposed GPM and LCM to OptimusVLA, we conduct extensive ablation studies on simulation benchmark and Real-World.

\noindent\textbf{Ablation on GPM and LCM}. As shown in Table \ref{tb:ablation}, removing GPM causes a catastrophic performance drop of 3.8\% on CALVIN and 9.4\% on the \textit{Generalization Tasks}. It can be attributed to the fact that, without GPM, OptimusVLA collapses into a standard flow-based policy whose  generalization is impaired by the large gap between the prior and target distributions. This highlights the crucial role of GPM in enhancing the cross-environment and cross-scene generalization of OptimusVLA. Similarly, removing LCM causes a 1.7\% drop on LIBERO-Long. This occurs because OptimusVLA loses the ability to generate temporally consistent actions. It highlights the critical role of LCM in maintaining trajectory smoothness and progress awareness.

\input{figures/fig-7}
\noindent\textbf{Ablation on Memory size}. As shown in Table \ref{tb:ab_memory}, we conduct experiments to investigate how the size of the memory bank in GPM affect overall performance. We observe that performance on LIBERO-Long scales with the richness of the memory bank. Storing only one trajectory per task causes performance to degrade, as the prior becomes too deterministic. Moreover, retrieving a sufficiently large $k$ (e.g., $k=8$) is crucial. A small $k$ overfits to a single retrieved trajectory, while a larger $k$ allows GPM to construct a robust Gaussian mixture prior, balancing specificity with exploratory potential.

\subsection{Efficiency Analysis}
\noindent\textbf{Training Efficiency}. As illustrated in Fig. \ref{fig:fig4}, we compare the performance of OptimusVLA and $\pi_{0.5}$ on LIBERO under different numbers of training steps. The results show that OptimusVLA achieves an 97.6\% success rate on LIBERO-Goal in just 18,000 steps, while $\pi_{0.5}$ requires 26,000 steps to reach a similar level. We attribute this acceleration to the task-level priors provided by GPM. Unlike standard approaches that map from uninformative isotropic noise to the action space, OptimusVLA initiates inference from retrieved priors. This strategy effectively places the initialization in the vicinity of the target manifold, thereby substantially reducing the complexity of the transformation.

\noindent\textbf{Inference Efficiency}. As illustrated in Fig. \ref{fig:fig5}, we compare the inference time and NFEs of OptimusVLA against $\pi_{0.5}$ on simulation benchmark and Real-World. OptimusVLA achieves 6.5$\times$ faster inference time and 3.1$\times$ fewer NFEs on LIBERO, while simultaneously delivering the highest task performance. We attribute these gains to the lightweight design of GPM and LCM, which introduce only minimal overhead yet substantially reduce the required NFEs—one of the primary factors governing inference speed.

\subsection{Qualitative Analysis} 
In Fig. \ref{fig:fig7}, we present qualitative examples from both simulation and real-world tasks. The top panel shows key frames of LIBERO task, \textit{Pick up the bbq sauce and place it in the basket}. In $\pi_{0.5}$, the prior distribution is Gaussian noise, and actions are sampled using a fixed NFEs. This introduces a potential risk that the sampled action is infeasible. In contrast, GPM enables OptimusVLA to start inference in the vicinity of the target action distribution, with adaptive noise scale and NFEs. This results in faster inference and improved performance. The bottom panel illustrates real-world case, \textit{Place the red apple onto the plate}. The $\pi_{0.5}$ lacks temporal awareness, thus, it struggles to determine if the apple has already been correctly placed when faced with visually similar observations. In contrast, LCM imposes a consistency constraint on OptimusVLA, leading to smoother generated trajectories and mitigating the adverse effects of similar observations. 


%% file: table/tb_main_robotwin.tex
\begin{table*}[h]
\centering
\caption{Performance comparison on RoboTwin 2.0 \cite{chen2025robotwin2}. We report per-task success rates (SR) and rank (RK) over 100 rollouts under \textit{Hard} setting. $^{\dag}$ represents the result we reproduced. 16 tasks results are available in Appendix.}
\label{tb:main_robotwin}
\small
\renewcommand{\arraystretch}{1.1}
\setlength{\tabcolsep}{3pt}

\begin{tabular*}{\textwidth}{@{}l|@{\extracolsep{\fill}}*{14}{c}@{}}
\toprule[1.2pt]
\multirow{2}{*}{\textbf{Task}}
& \multicolumn{2}{c}{\textbf{RDT} \cite{liu2024rdt}}
& \multicolumn{2}{c}{\textbf{ACT} \cite{zhao2023act}}
& \multicolumn{2}{c}{\textbf{DP} \cite{chi2025diffusionpolicy}}
& \multicolumn{2}{c}{\textbf{DP3} \cite{ze2024dp3}}
& \multicolumn{2}{c}{$\mathbf{\pi_{0}}$ \cite{black2024pi0}}
& \multicolumn{2}{c}{$\mathbf{\pi_{0.5}}^{\dag}$\cite{intelligence2025pi05}}
& \multicolumn{2}{>{\columncolor[HTML]{E7EEFE}}c}{\textbf{OptimusVLA}} \\
\cmidrule(lr){2-3}\cmidrule(lr){4-5}\cmidrule(lr){6-7}\cmidrule(lr){8-9}%
\cmidrule(lr){10-11}\cmidrule(lr){12-13}\cmidrule(lr){14-15}
& SR & RK
& SR & RK
& SR & RK
& SR & RK
& SR & RK
& SR & RK
& \multicolumn{2}{>{\columncolor[HTML]{E7EEFE}}c}{%
    \begin{tabular}{@{}cc@{}}
        SR & RK
    \end{tabular}
} \\
\midrule
Click Alarmclock
& 12\% & 4
& 4\%  & 7
& 5\%  & 6
& 14\% & 3
& 11\% & 5
& 18\% & 2
& \multicolumn{2}{>{\columncolor[HTML]{E7EEFE}}c}{%
    \begin{tabular}{@{}cc@{}}
        31\% & 1
    \end{tabular}
} \\
Click Bell
& 9\% & 3
& 3\% & 4
& 0\% & 5
& 0\% & 5
& 3\% & 4
& 28\% & 2
& \multicolumn{2}{>{\columncolor[HTML]{E7EEFE}}c}{%
    \begin{tabular}{@{}cc@{}}
        46\% & 1
    \end{tabular}
} \\
Dump Bin Bigbin
& 32\% & 3
& 1\%  & 6
& 0\%  & 7
& 53\% & 1
& 24\% & 5
& 29\% & 4
& \multicolumn{2}{>{\columncolor[HTML]{E7EEFE}}c}{%
    \begin{tabular}{@{}cc@{}}
        35\% & 2
    \end{tabular}
} \\
Open Laptop
& 32\% & 4
& 0\% & 6
& 0\% & 6
& 7\% & 5
& 46\% & 2
& 38\% & 3
& \multicolumn{2}{>{\columncolor[HTML]{E7EEFE}}c}{%
    \begin{tabular}{@{}cc@{}}
        48\% & 1
    \end{tabular}
} \\
Place Bread Skillet
& 1\% & 3
& 0\% & 4
& 0\% & 4
& 0\% & 4
& 1\% & 4
& 2\% & 2
& \multicolumn{2}{>{\columncolor[HTML]{E7EEFE}}c}{%
    \begin{tabular}{@{}cc@{}}
        4\% & \; 1
    \end{tabular}
} \\
Place Container Plate
& 17\% & 4
& 1\%  & 5
& 0\%  & 6
& 1\%  & 5
& 45\% & 1
& 30\% & 3
& \multicolumn{2}{>{\columncolor[HTML]{E7EEFE}}c}{%
    \begin{tabular}{@{}cc@{}}
        37\% & 2
    \end{tabular}
} \\
Press Stapler
& 24\% & 4
& 6\%  & 5
& 0\%  & 7
& 3\%  & 6
& 29\% & 3
& 36\% & 2
& \multicolumn{2}{>{\columncolor[HTML]{E7EEFE}}c}{%
    \begin{tabular}{@{}cc@{}}
        45\% & 1
    \end{tabular}
} \\
Stack Bowls Two
& 30\% & 4
& 0\% & 6
& 0\% & 6
& 6\% & 5
& 41\% & 3
& 49\% & 2
& \multicolumn{2}{>{\columncolor[HTML]{E7EEFE}}c}{%
    \begin{tabular}{@{}cc@{}}
        58\% & 1
    \end{tabular}
} \\
\hline
Average
& 20\% & 4
& 2\%  & 6
& 1\%  & 7
& 11\% & 5
& 25\% & 3
& 29\% & 2
& \multicolumn{2}{>{\columncolor[HTML]{E7EEFE}}c}{%
    \begin{tabular}{@{}cc@{}}
        38\% & 1
    \end{tabular}
} \\
\bottomrule[1.2pt]
\end{tabular*}
\end{table*}

%% file: table/tb_ablation.tex
\begin{table}[t]
\centering
\caption{Ablation study of GPM and LCM on LIBERO-Long, CALVIN, and Real-World \textit{Generalization Tasks}.}
\label{tb:ablation}
\renewcommand\arraystretch{1.1}
\resizebox{0.5\textwidth}{!}{%
\begin{tabular}{cc|ccc}
\toprule[1.1pt]
\multicolumn{2}{c|}{\textbf{Ablation Setting}}                                                                         & \multicolumn{2}{c}{\textbf{Simulation}}  & \textbf{Real-World}       \\ \hline
GPM          & LCM      & LIBERO-Long & CALVIN &  \textit{Generalization Tasks}  \\ \hline
\rowcolor[HTML]{E7EEFE}
\Checkmark   &       \Checkmark & \textbf{96.4} & \textbf{4.45} & \textbf{85.0}   \\
  & \Checkmark &  93.2 \textcolor{red}{\scriptsize ($\downarrow$ 3.3\%)}    & 4.28 \textcolor{red}{\scriptsize ($\downarrow$ 3.8\%)}    & 77.0 \textcolor{red}{\scriptsize ($\downarrow$ 9.4\%)}  \\
 \Checkmark   &       & 94.8 \textcolor{red}{\scriptsize ($\downarrow$ 1.7\%)}   & 4.38 \textcolor{red}{\scriptsize ($\downarrow$ 1.6\%)}  &   79.5 \textcolor{red}{\scriptsize ($\downarrow$ 6.5\%)}  \\
    &          &  92.4 \textcolor{red}{\scriptsize ($\downarrow$ 4.1\%)}      & 4.26  \textcolor{red}{\scriptsize ($\downarrow$ 4.3\%)}  &   75.0 \textcolor{red}{\scriptsize ($\downarrow$ 11.8\%)}   \\
\bottomrule[1.1pt]
\end{tabular}
}
\vspace{-1em}
\end{table}

%% file: table/tb_ab_memory.tex
\begin{table}[htbp]
\centering
\caption{Ablation study on memory size of GPM. We report average success rate on LIBERO-Long. \texttt{Num}. denotes the number of trajectories in the memory bank, \texttt{k}. denotes the number of retrieved trajectories.}
\label{tb:ab_memory}
\small
\begin{tabular}{l|ccc|cc|c}
\toprule[1.1pt]
\multirow{2}{*}{Metric} & \multicolumn{3}{c|}{\texttt{Num}=6500} & \multicolumn{2}{c|}{\texttt{Num}=1300} & \multicolumn{1}{c}{\texttt{Num}=130} \\
 & $k{=}8$ & $k{=}16$ & $k{=}1$ & $k{=}8$ & $k{=}1$ & $k{=}8$ \\
\midrule
SR       & \textbf{96.4} & 94.8 & 92.6 & 95.2 & 92.4 & 93.6 \\

\bottomrule[1.1pt]
\end{tabular}
\vspace{-3pt}
\end{table}

%% file: figures/fig-3.tex
\begin{figure*}[htbp]
    \centering
    \includegraphics[width=1\textwidth]{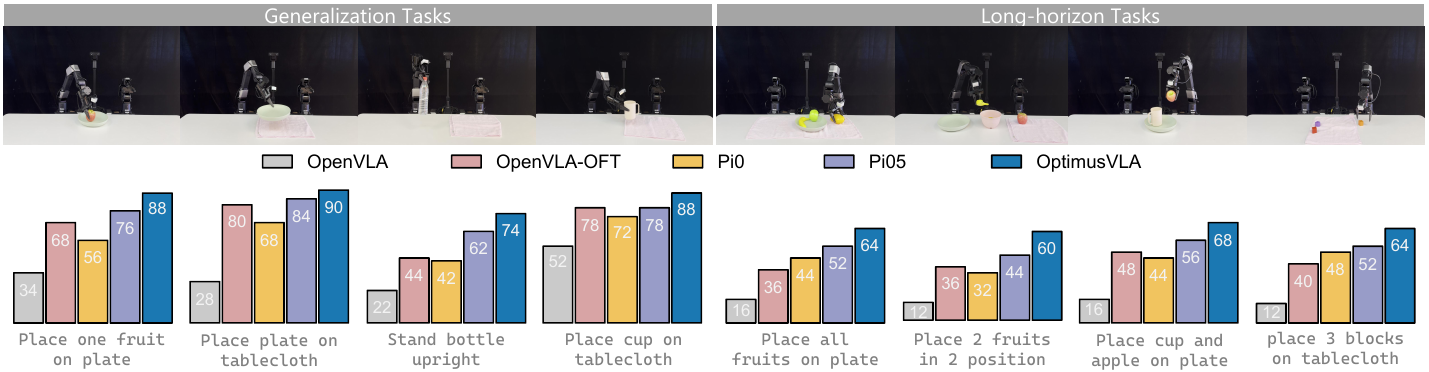}
    \caption{Real-world task setup and evaluation results. We evaluate the performance of OptimusVLA against OpenVLA \cite{kim2024openvla}, OpenVLA-OFT \cite{kim2025openvla-oft}, $\pi_{0}$ \cite{black2024pi0}, and $\pi_{0.5}$ \cite{intelligence2025pi05} on the \textit{Generalization Tasks} and \textit{Long-horizon Tasks} suites.}
    \label{fig:fig3}
\end{figure*}

%% file: figures/fig-4.tex
\begin{figure}[htbp]
    \centering
    \begin{subfigure}[b]{0.48\linewidth}
    \centering
    \includegraphics[width=\linewidth]{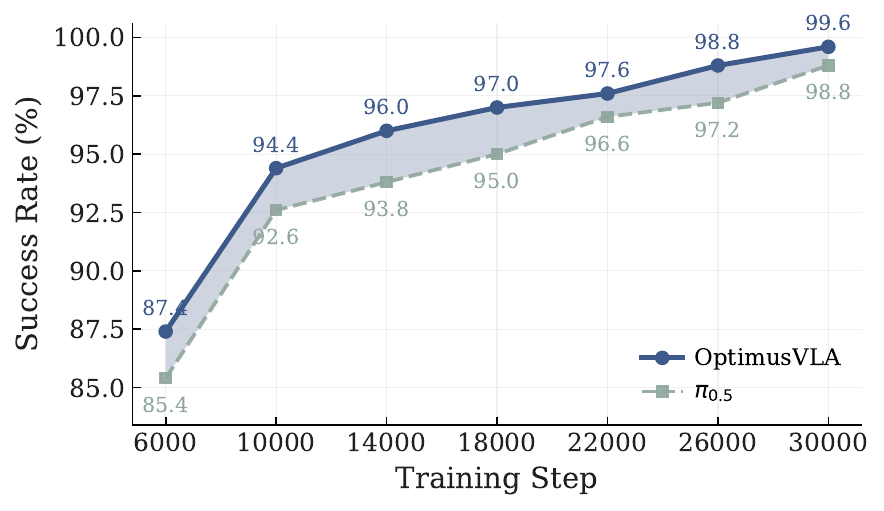}
    \caption{LIBERO-Spatial}
    \label{fig:exa:a}
  \end{subfigure}\hfill
  \begin{subfigure}[b]{0.48\linewidth}
    \centering
    \includegraphics[width=\linewidth]{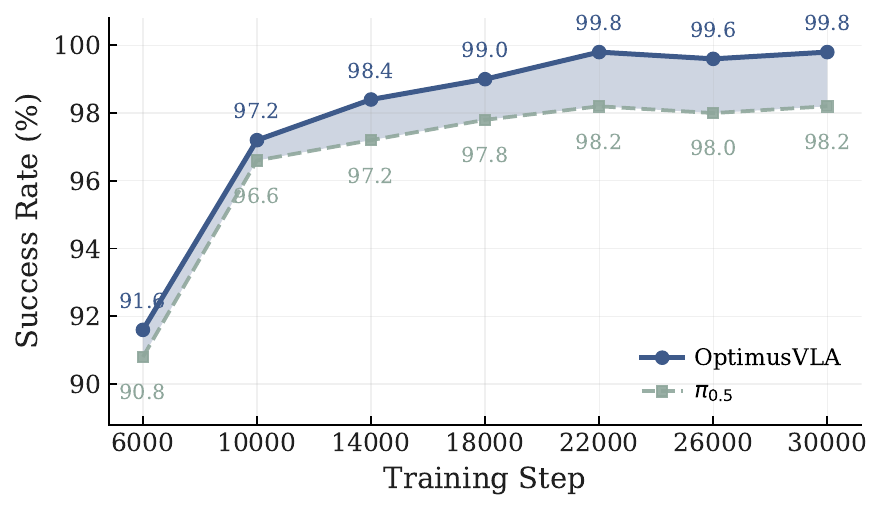}
    \caption{LIBERO-Object}
    \label{fig:exa:b}
  \end{subfigure}

  \vspace{0.6em}

  \begin{subfigure}[b]{0.48\linewidth}
    \centering
    \includegraphics[width=\linewidth]{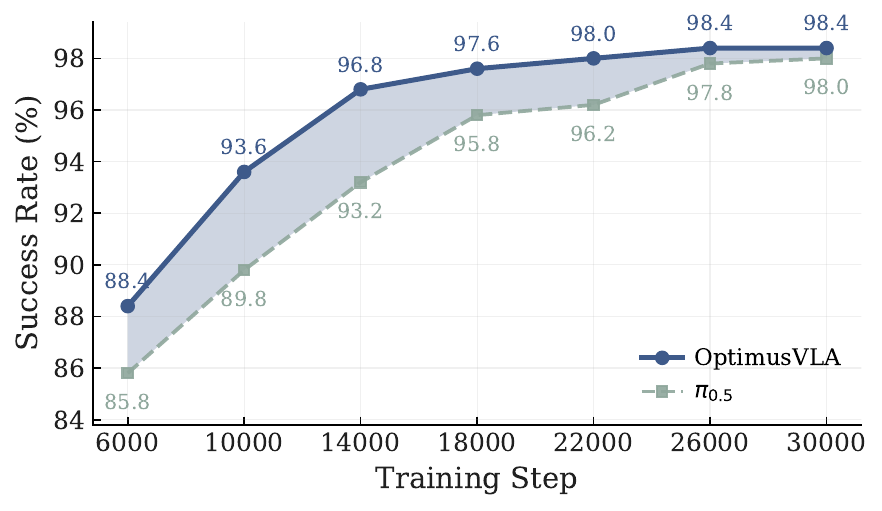}
    \caption{LIBERO-Goal}
    \label{fig:exa:c}
  \end{subfigure}\hfill
  \begin{subfigure}[b]{0.48\linewidth}
    \centering
    \includegraphics[width=\linewidth]{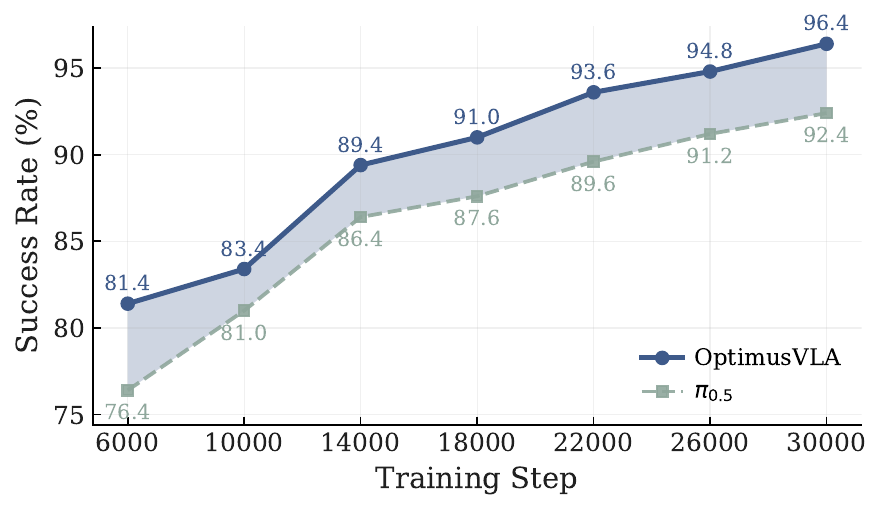}
    \caption{LIBERO-Long}
    \label{fig:exa:d}
  \end{subfigure}
    \caption{Training efficiency comparison between OptimusVLA and 
$\pi_{0.5}$ \cite{intelligence2025pi05}. Initialized from same weights, OptimusVLA attains strong performance with substantially fewer training steps.}
    \label{fig:fig4}
    \vspace{-1em}
\end{figure}

%% file: figures/fig-5.tex

\begin{figure}[htbp]
    \centering
    \includegraphics[width=0.5\textwidth]{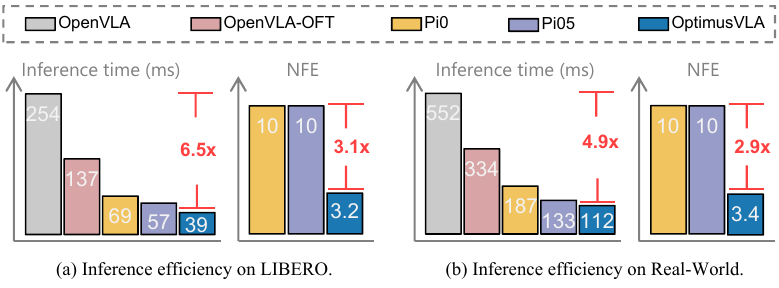}
    \vspace{-7pt}
    \caption{Inference efficiency comparison on LIBERO and Real-World. OptimusVLA attains strong performance with substantially fewer inference time and NFEs.}
    \label{fig:fig5}
    \vspace{-11pt}
\end{figure}

%% file: figures/fig-7.tex
\begin{figure*}[htbp]
    \centering
    \includegraphics[width=1\textwidth]{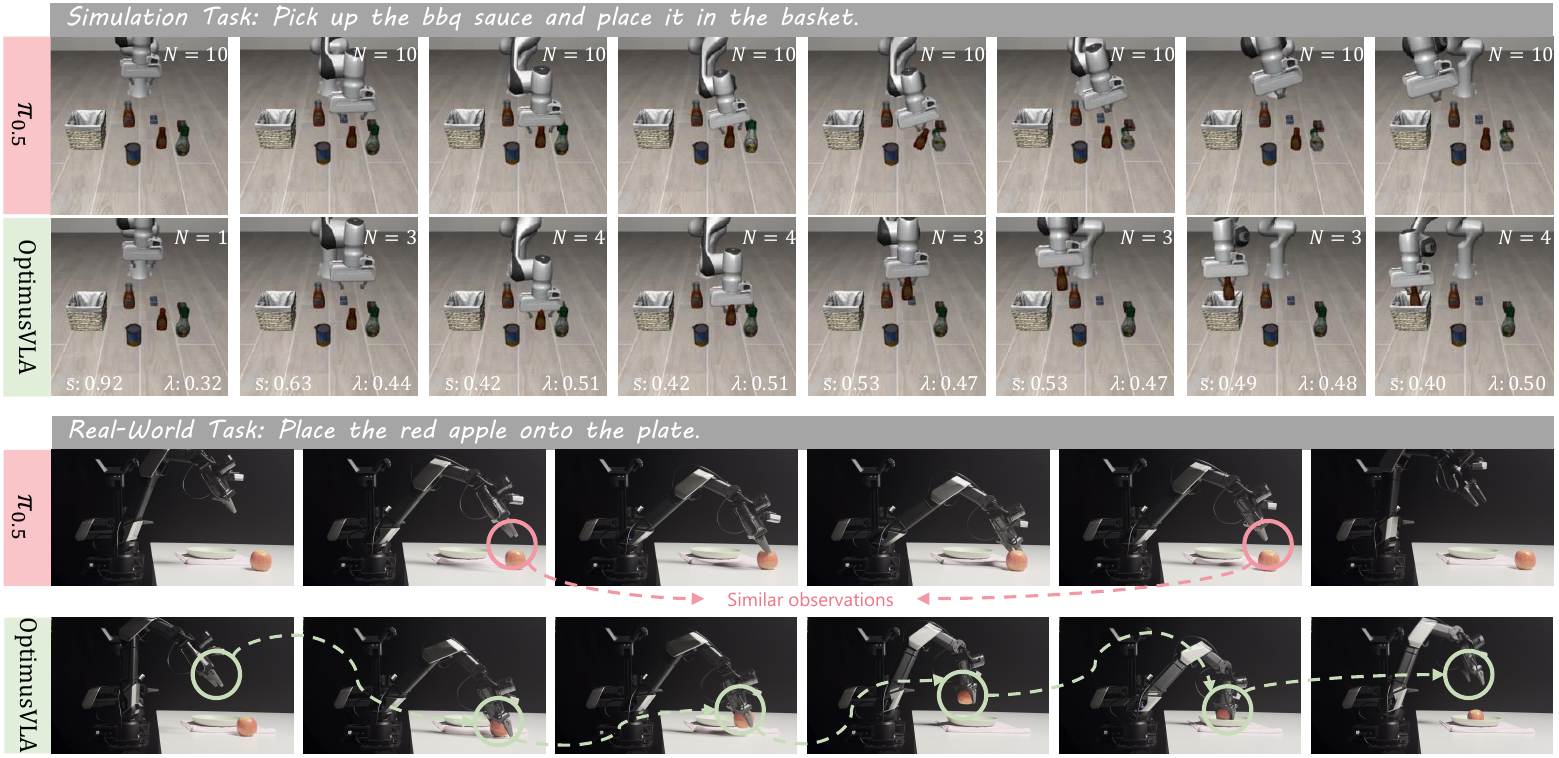}
    \caption{Qualitative results of OptimusVLA on simulation task and Real-World task. \textbf{Top}: On simulation task, we visualize the retrieval similarity $\bar s$, adaptive noise scales $\lambda$, and NFEs $N$ of key frames. \textbf{Bottom}: On Real-World task, we demonstrate the role of LCM in modeling temporal dependencies, whereas $\pi_{0.5}$ struggles to distinguish between similar observations.}
    \label{fig:fig7}
    \vspace{-8pt}
\end{figure*}

%% file: sec/5_conclusion.tex
\section{Conclusion}
In this paper, we propose a dual-memory VLA framework, OptimusVLA, which contain Global Prior Memory (GPM) and Local Consistency Memory (LCM) for robotic manipulation. GPM replaces Gaussian noise with task-level priors retrieved from semantically similar trajectories, thereby shortening the generative path and reducing invalid samples without sacrificing generalization. LCM models short histories of executed actions to infer task progress and injects a learned consistency constraint that enforces temporal coherence. Together, GPM and LCM improve efficiency and robustness of OptimusVLA. Extensive experiments in both simulation platforms and the real world demonstrate the superior performance of OptimusVLA, together with substantially higher inference efficiency.

%% file: Supplementary/X_suppl.tex
\clearpage
\setcounter{page}{1}
\setcounter{section}{0}
\maketitlesupplementary
\renewcommand{\thesection}{\Alph{section}}

\noindent The supplementary document is organized as follows:

\begin{itemize}
\setlength{\itemsep}{10pt}
    \item Sec. \ref{sup:limitation}: Limitation and Future Work.
    
    \item Sec. \ref{sup:arc}: Global Prior Memory.
    
    \item Sec. \ref{sup:train}: Training Details.
    
    \item Sec. \ref{sup:eval}: Evaluation.
    
    \item Sec. \ref{sup:case}: Case Study.
\end{itemize}

\section{Limitation and Future Work}
\label{sup:limitation}
Although OptimusVLA improves both efficiency and robustness of VLA model through Global Prior Memory and Local Consistency Memory, several limitations remain. First, the effectiveness of GPM is inherently constrained by the coverage and quality of the trajectory memory bank. When the current task or scene substantially departs from stored experiences, the retrieved priors may be misleading, potentially biasing the VLA model toward suboptimal behaviors. Second, LCM focuses on local, action-centric temporal coherence and operates on fixed-length chunks. While this design keeps the overhead small, it may be insufficient for tasks that require reasoning over very long horizons, multi-stage dependencies, or delayed effects. 

A natural direction is to develop adaptive memory mechanisms, where the GPM bank is updated online with consolidation, forgetting, and uncertainty-aware retrieval, enabling continual learning and more robust behavior under distribution shift. Another avenue is to jointly train GPM, LCM, and the flow policy end-to-end. These are remains as future work for this paper.

\section{Global Prior Memory}
\label{sup:arc}
\subsection{Preliminaries}

We begin from Conditional Flow Matching (CFM) for action generation. Let $x_0 \sim \mathcal{P}_0$ be a source sample (typically noise) and $x_1 \sim \mathcal{P}_1$ an action from the data distribution. CFM defines a straight-line probability path between distributions:
\begin{equation}
x_t = (1-t)x_0 + t x_1, \qquad t \in [0,1],
\end{equation}
implying a constant target velocity field $u_t(x_t \mid x_0,x_1) = x_1 - x_0$. The flow policy $v_\theta(t,x)$ is trained to regress this vector field by minimizing:
\begin{equation}
\mathcal{L}_{\mathrm{CFM}}(\theta) = \mathbb{E}_{t\sim\mathcal{U}[0,1],\,x_0\sim\mathcal{P}_0,\,x_1\sim\mathcal{P}_1}
\big\| v_\theta(t,x_t) - (x_1 - x_0) \big\|_2^2.
\label{eq:cfm_app}
\end{equation}
At inference, actions are generated by solving the Ordinary Differential Equation (ODE) $dx_t/dt = v_\theta(t,x_t)$.
Standard approaches set the source distribution $\mathcal{P}_0 = \mathcal{N}(0, I)$ during both training and inference. However, mapping isotropic Gaussian noise to the complex manifold of robotic actions requires a high Number of Function Evaluations (NFE).
In OptimusVLA, we maintain $\mathcal{P}_0 = \mathcal{N}(0, I)$ during the pre-training of the flow policy to learn a generalizable vector field. Crucially, at inference time, GPM intervenes in this initialization stage. We construct a task-level prior $\mathcal{P}_{\mathrm{re}}$ from retrieved memories and replace the standard noise source with $\mathcal{P}_{\mathrm{re}}$. This places the starting point of the flow ODE significantly closer to the target data manifold, enabling efficient generation with reduced NFE.

\subsection{Offline Construction}
Let $\mathcal{D} = \{(O^{(m)}, \ell^{(m)}, J^{(m)})\}_{m=1}^M$ be a dataset of episodes, where $J^{(m)} = \{a^{(m)}_0,\dots,a^{(m)}_{T_m-1}\}$ is the dense action sequence. We employ the pre-trained VLM backbone to extract a compact representation for each episode. Specifically, we perform a forward pass to obtain multimodal prefix tokens $H^{(m)} \in \mathbb{R}^{L \times D_{\mathrm{vlm}}}$ and compute a mean-pooled embedding $e^{(m)}$:
\begin{equation}
e^{(m)} = \frac{1}{L} \sum_{j=1}^{L} H^{(m)}_{j} \in \mathbb{R}^{D_{\mathrm{vlm}}}.
\end{equation}
This embedding is projected by the Prior Head $f_{\mathrm{head}}$ (a 2-layer MLP) into a normalized task embedding:
\begin{equation}
z_m = f_{\mathrm{head}}(e^{(m)}), \qquad
\tilde z_m = \frac{z_m}{\|z_m\|_2} \in \mathbb{R}^{D_z}.
\end{equation}
The Prior Head is trained via an InfoNCE objective to ensure that embeddings from the same semantic task are clustered together while distinct tasks are separated. Formally, for a query $z_i$, the loss is:
\begin{equation}
\mathcal{L}_{\mathrm{InfoNCE}} = -\log \frac{\sum_{j \in \mathcal{P}(i), j \neq i} \exp(\langle \tilde z_i, \tilde z_j \rangle / \tau_c)}{\sum_{k \in \mathcal{A}(i), k \neq i} \exp(\langle \tilde z_i, \tilde z_k \rangle / \tau_c)},
\label{eq:infonce_app}
\end{equation}
where $\mathcal{P}(i)$ is the set of positive indices (same task) and $\mathcal{A}(i)$ includes all indices in the batch.
After training, we freeze $f_{\mathrm{head}}$ and construct a Key-Value Memory Bank:
\begin{equation}
\mathcal{M} = \big\{ (\tilde z_m, J^{(m)}, \text{meta}^{(m)}) \big\}_{m=1}^{M},
\end{equation}
where $\text{meta}^{(m)}$ contains trajectory metadata (length, chunking parameters). For efficient retrieval, we build a FAISS \texttt{IndexFlatIP} index over the keys $\{\tilde z_m\}$.

\subsection{Online Retrieval and Prior Construction}
At inference time $t$, given observation $O_t$ and instruction $\ell$, the VLM outputs current tokens $H_t$. We compute the query token $\tilde z_{re}$ similarly:
\begin{equation}
e_t = \frac{1}{L} \sum_{j=1}^{L} H_{t,j}, \qquad \tilde z_{re} = \frac{f_{\mathrm{head}}(e_t)}{\|f_{\mathrm{head}}(e_t)\|_2}.
\end{equation}
Querying the memory bank yields the top-$k$ nearest trajectories with cosine similarity scores $s_i = \langle \tilde z_{re}, \tilde z_i \rangle$. We compute soft weights $\alpha_i$ and a global similarity metric $\bar{s}$:
\begin{equation}
\alpha_i = \frac{\exp(s_i / \tau_s)}{\sum_{j=1}^k \exp(s_j / \tau_s)}, \qquad
\bar s = \sum_{i=1}^{k} \alpha_i s_i.
\end{equation}
Here, $\bar s \in [-1,1]$ serves as a confidence indicator for the adaptive scheduler.

To construct the action prior, we align the retrieved full trajectories $J_i$ to the current execution progress. We maintain a progress scalar $\rho_t \in [0,1]$ and extract the relevant action chunk $C_i$ from each neighbor $J_i$. Using sliding windows of length $H_0$ and stride $\Delta$, we identify the window index $u_i = \lfloor \rho_t \cdot (N_{\mathrm{chunks}}^{(i)}-1) \rfloor$ and extract:
\begin{equation}
C_i = \mathrm{Resample}\big( J_i[u_i \cdot \Delta : u_i \cdot \Delta + H_0] \big) \in \mathbb{R}^{H \times A}.
\end{equation}
If the retrieved chunk length $H_0$ differs from the model's horizon $H$, we perform linear interpolation to match dimensions.
The task-level prior $\mathcal{P}_{\mathrm{re}}$ is then modeled as a Gaussian Mixture approximated by a single Moment-Matched Gaussian $\mathcal{N}(\mu, \Sigma)$. The mean $\mu$ and diagonal covariance $\Sigma = \mathrm{diag}(\mathrm{Var})$ are:
\begin{equation}
\mu = \sum_{i=1}^{k} \alpha_i C_i, \qquad
\mathrm{Var} = \sum_{i=1}^{k} \alpha_i (C_i - \mu)^{\odot 2},
\label{eq:prior_stats}
\end{equation}
where operations are element-wise. We enforce a minimum variance floor $\sigma_{\min}^2$ to prevent collapse.

\input{Supplementary/tb_parameters}
\subsection{Similarity-Adaptive Sampling}
GPM dynamically adjusts the generative process based on retrieval confidence $\bar s$. We define monotonic mappings for the noise scale $\lambda$ and discretization steps $N$:
\begin{equation}
\lambda(\bar s) = \lambda_{\max} - \frac{\bar s+1}{2}(\lambda_{\max} - \lambda_{\min}),
\end{equation}
\begin{equation}
N(\bar s) = \mathrm{Round}\left( N_{\min} + \left(1 - \frac{\bar s+1}{2}\right)(N_{\max} - N_{\min}) \right).
\end{equation}
When retrieval confidence is high ($\bar s \approx 1$), $\lambda$ decreases (relying more on the retrieved mean) and $N$ decreases (easier transport); conversely, for novel scenarios ($\bar s \approx 0$), the model gracefully falls back to higher noise and more compute steps. The initialization $\mathbf{\hat{X}}_t$ for the flow policy is sampled as:
\begin{equation}
\epsilon \sim \mathcal{N}(0, I), \qquad
\mathbf{\hat{X}}_t = \mu + \lambda(\bar s) \cdot \big( \epsilon \odot \sqrt{\mathrm{Var}} \big).
\end{equation}
This $\mathbf{\hat{X}}_t$ replaces the standard Gaussian noise $x_0$ in Eq. \ref{eq:cfm_app}.

\subsection{Session-Level Caching}
To minimize latency, we implement a \texttt{MemorySession} module. Retrieval is performed once at the start of an episode (or when significant task drift is detected). The top-$k$ trajectory indices and weights are cached. At each subsequent step $t$, the session efficiently gathers the time-aligned chunks $C_i$ based on $\rho_t$ and computes Eq. \ref{eq:prior_stats} without re-querying the FAISS index, ensuring negligible overhead.

\section{Training Details}
\label{sup:train}
We now provide a more detailed description of the three-stage training pipeline used to obtain OptimusVLA. The hyperparameter settings are shown in Table \ref{tab:hyperparams}.
\input{Supplementary/tb_robotwin_all}
\subsection{Stage I: Hierarchical VLA Pre-training}
\label{app:stage1}

In the first stage, we pretrain a hierarchical Vision–Language–Action model following the architecture and training protocol of $\pi_{0.5}$~\cite{intelligence2025pi05}. At this point, neither Global Prior Memory (GPM) nor Local Consistency Memory (LCM) is attached; the goal is to obtain a strong base VLA that serves as the backbone for subsequent stages.

Given an instruction $\ell$ and observation $O_t$, the Vision–Language backbone \texttt{VLM} produces a multimodal representation $E_{emb} = \texttt{VLM}(O_t, \ell)$, which is fed into the flow policy. We extract ground-truth action chunks $\mathbf{A}^\star_t \in \mathbb{R}^{H \times A}$ from the dataset and train the flow policy $p_{\theta}$ using a Conditional Flow Matching (CFM) objective.

\subsection{Stage II: GPM Training}
\label{app:stage2}

In the second stage, we attach a lightweight Prior Head on top of the pretrained Vision–Language backbone. All parameters of the base VLA model are frozen, and only the Prior Head is updated.

\noindent\textbf{Training Procedure}.
We train the Prior Head to minimize the task-contrastive loss $\mathcal{L}_{\mathrm{InfoNCE}}$ (Eq. \ref{eq:infonce_app}).
Crucially, to ensure the objective is computable and robust, we employ a Task-Pair Batch Sampler. Unlike standard random sampling, this sampler groups the training data by task ID and constructs mini-batches such that each batch contains at least two trajectories from the same task.

This strictly batched sampling strategy guarantees that every anchor sample has at least one positive pair within the current batch (in-batch positives), while all trajectories from other tasks serve as hard negatives. We optimize the head using AdamW with a learning rate of $1\times 10^{-4}$, a batch size of 64, and a temperature $\tau_c=0.07$ for 20 epochs.

\subsection{Stage III: LCM Training}
\label{app:stage3}

In the final stage, we train the LCM. The VLM backbone, flow policy, and GPM are all frozen.

\noindent\textbf{Training Targets}.
LCM is trained to predict the residual needed to bridge the gap between the global prior and the ground truth. For the time step $t$, we first retrieve the top-$k$ priors using GPM and compute the Gaussian mean $\mu_t$ (as defined in Eq.~\eqref{eq:prior_stats}). The regression target is:
\begin{equation}
   \mathbf{B}^\star_t = \mathbf{A}^\star_t - \mu_t. 
\end{equation}

\noindent\textbf{Optimization}.
We unroll the LCM model along the trajectory. The model takes the previous action chunk $\mathbf{A}_{t-1}$ as input. To ensure robustness during the initial inference steps (where no history exists), we employ a cold-start strategy during training: with probability $p_{cold}$, we mask $\mathbf{A}_{t-1}$ with zeros, forcing the model to rely solely on internal dynamics or output a neutral bias. The loss is the MSE between the predicted bias $\mathbf{B}_t$ and the target $\mathbf{B}^\star_t$.

\input{Supplementary/tb_real_world_generazation}
\input{Supplementary/tb_real_world_long}
\section{Evaluation}
\label{sup:eval}

\subsection{Evaluation on RoboTwin 2.0}

RoboTwin~2.0 defines a standardized bimanual manipulation benchmark built on the RoboTwin-OD object library, which contains $731$ annotated objects from $147$ categories, and a pre-collected dataset of over $100\text{k}$ expert dual-arm trajectories. The benchmark spans $50$ collaborative dual-arm tasks instantiated on five distinct robot embodiments. For the main simulation protocol, each task is trained in a single-task manner on the Aloha–AgileX dual-arm platform using $50$ clean expert demonstrations, and VLA models are tested with $100$ rollouts under two difficulty settings: an \emph{Easy} regime with uncluttered, clean scenes and a \emph{Hard} regime with strong domain randomization over clutter, background textures, lighting, and tabletop height. In this paper, we randomly select 16 tasks for evaluation, and test the models with 100 rollouts under the \emph{Hard} (domain-randomized with clutter, lighting, textures, and height variations) settings. As shown in Table \ref{tab:robotwin_all}, OptimusVLA achieves an average success rate of 30\% across 16 tasks, surpassing all baselines, including $\pi_{0.5}$.

\subsection{Evaluation on Real-World}
In the Real-World evaluation, we employ Galaxea R1 Lite as our robot platform. The Galaxea R1 Lite is a mobile, wheeled humanoid robot with a bimanual upper body, specifically designed for operation in human-centric indoor environments. Its embodiment comprises 23 degrees of freedom: two 6-DoF arms with spherical wrists and parallel two-finger grippers, a 3-DoF torso providing vertical and pitch motion for workspace extension, and a 6-DoF vector-drive omnidirectional base enabling coordinated whole-body manipulation.

As shown in Table \ref{tab:real-world-generalization} and Table \ref{tab:real-world-long-horizon}, we construct two task suites: \textit{Generalization Tasks} and \textit{Long-horizon Tasks}. The \textit{Generalization Tasks} primarily evaluate the model’s ability to generalize across varying scenes, lighting conditions, and object instances, whereas the \textit{Long-horizon Tasks} focus on assessing the stability and robustness of the model when executing long-horizon sequences. For each evaluation episode, we randomly initialize the objects at different spatial locations. The experimental results in  Table \ref{tab:real-world-generalization} and Table \ref{tab:real-world-long-horizon} demonstrate the superior performance of OptimusVLA in real-world environments. We show more visualization examples in the next section.
\section{Case Study}
\label{sup:case}
In this section, we present qualitative Real-World examples of OptimusVLA. Empowered by GPM and LCM, OptimusVLA exhibits superior performance on the \textit{Generalization Tasks} (Figs. \ref{fig:appendix-case1}) as well as on the \textit{Long-horizon Tasks} (Figs. \ref{fig:appendix-case2}). Moreover, we provide \textbf{unedited}, \textbf{real-time} video examples of OptimusVLA performing Real-World tasks in the supplementary materials (see \texttt{appendix/video}). Specifically, \textit{Place plate on tablecloth} evaluates the generalization ability of OptimusVLA to varying lighting conditions and object positions, while \textit{Place all fruits on plate} assesses the stability of OptimusVLA when executing long-horizon action sequences.

\input{Supplementary/fig-case1}
\input{Supplementary/fig-case2}

%% file: Supplementary/tb_parameters.tex
\begin{table}[ht]
\centering
\caption{Hyperparameter setting for each training phase.}
\label{tab:hyperparams}
\renewcommand\arraystretch{1.1}
\begin{tabular}{lccc}
\toprule[1.2pt]
Hyperparameter & Stage-1 & Stage-2 & Stage-3 \\ \hline
Optimizer      & AdamW   & AdamW      & AdamW       \\
Learning Rate  & 5e-5  & 1e-4       & 1e-4     \\
Steps          & 30000   & 1000       & 1000          \\
Batch Size     & 512     & 64         & 64        \\
Warm Up Ratio  & 0.10    & -          & - \\
Ema Decay      & 0.999   & -          & - \\
\bottomrule[1.2pt]
\end{tabular}
\end{table}

%% file: Supplementary/tb_robotwin_all.tex
\begin{table*}[h]
\centering
\caption{Performance comparison on RoboTwin 2.0 \cite{chen2025robotwin2}. We report per-task success rates (SR) and rank (RK) over 100 rollouts under \textit{Hard} setting. $^{\dag}$ represents the result we reproduced.}
\label{tab:robotwin_all}
\small
\renewcommand{\arraystretch}{1.1}
\setlength{\tabcolsep}{3pt}

\begin{tabular*}{\textwidth}{@{}l|@{\extracolsep{\fill}}*{14}{c}@{}}
\toprule[1.2pt]
\multirow{2}{*}{\textbf{Task}}
& \multicolumn{2}{c}{\textbf{RDT} \cite{liu2024rdt}}
& \multicolumn{2}{c}{\textbf{ACT} \cite{zhao2023act}}
& \multicolumn{2}{c}{\textbf{DP} \cite{chi2025diffusionpolicy}}
& \multicolumn{2}{c}{\textbf{DP3} \cite{ze2024dp3}}
& \multicolumn{2}{c}{$\mathbf{\pi_{0}}$ \cite{black2024pi0}}
& \multicolumn{2}{c}{$\mathbf{\pi_{0.5}}^{\dag}$\cite{intelligence2025pi05}}
& \multicolumn{2}{>{\columncolor[HTML]{E7EEFE}}c}{\textbf{OptimusVLA}} \\
\cmidrule(lr){2-3}\cmidrule(lr){4-5}\cmidrule(lr){6-7}\cmidrule(lr){8-9}%
\cmidrule(lr){10-11}\cmidrule(lr){12-13}\cmidrule(lr){14-15}
& SR & RK
& SR & RK
& SR & RK
& SR & RK
& SR & RK
& SR & RK
& \multicolumn{2}{>{\columncolor[HTML]{E7EEFE}}c}{%
    \begin{tabular}{@{}cc@{}}
        SR & RK
    \end{tabular}
} \\
\midrule
Click Alarmclock
& 12\% & 4
& 4\%  & 7
& 5\%  & 6
& 14\% & 3
& 11\% & 5
& 18\% & 2
& \multicolumn{2}{>{\columncolor[HTML]{E7EEFE}}c}{%
    \begin{tabular}{@{}cc@{}}
        31\% & 1
    \end{tabular}
} \\
Click Bell
& 9\% & 3
& 3\% & 4
& 0\% & 5
& 0\% & 5
& 3\% & 4
& 28\% & 2
& \multicolumn{2}{>{\columncolor[HTML]{E7EEFE}}c}{%
    \begin{tabular}{@{}cc@{}}
        46\% & 1
    \end{tabular}
} \\
Dump Bin Bigbin
& 32\% & 3
& 1\%  & 6
& 0\%  & 7
& 53\% & 1
& 24\% & 5
& 29\% & 4
& \multicolumn{2}{>{\columncolor[HTML]{E7EEFE}}c}{%
    \begin{tabular}{@{}cc@{}}
        35\% & 2
    \end{tabular}
} \\
Open Laptop
& 32\% & 4
& 0\% & 6
& 0\% & 6
& 7\% & 5
& 46\% & 2
& 38\% & 3
& \multicolumn{2}{>{\columncolor[HTML]{E7EEFE}}c}{%
    \begin{tabular}{@{}cc@{}}
        48\% & 1
    \end{tabular}
} \\
Place Bread Skillet
& 1\% & 3
& 0\% & 4
& 0\% & 4
& 0\% & 4
& 1\% & 4
& 2\% & 2
& \multicolumn{2}{>{\columncolor[HTML]{E7EEFE}}c}{%
    \begin{tabular}{@{}cc@{}}
        4\% & \ 1
    \end{tabular}
} \\
Place Container Plate
& 17\% & 4
& 1\%  & 5
& 0\%  & 6
& 1\%  & 5
& 45\% & 1
& 30\% & 3
& \multicolumn{2}{>{\columncolor[HTML]{E7EEFE}}c}{%
    \begin{tabular}{@{}cc@{}}
        37\% & 2
    \end{tabular}
} \\
Press Stapler
& 24\% & 4
& 6\%  & 5
& 0\%  & 7
& 3\%  & 6
& 29\% & 3
& 36\% & 2
& \multicolumn{2}{>{\columncolor[HTML]{E7EEFE}}c}{%
    \begin{tabular}{@{}cc@{}}
        45\% & 1
    \end{tabular}
} \\
Stack Bowls Two
& 30\% & 4
& 0\% & 6
& 0\% & 6
& 6\% & 5
& 41\% & 3
& 49\% & 2
& \multicolumn{2}{>{\columncolor[HTML]{E7EEFE}}c}{%
    \begin{tabular}{@{}cc@{}}
        58\% & 1
    \end{tabular}
} \\
Beat Block Hammer
& 37\% & 1 
& 3\%  & 5
& 0\%  & 6
& 8\% & 4
& 21\% & 3
& 21\% & 3
& \multicolumn{2}{>{\columncolor[HTML]{E7EEFE}}c}{%
    \begin{tabular}{@{}cc@{}}
        26\% & 2
    \end{tabular}
} \\
Lift Pot
& 9\% & 4
& 0\% & 5
& 0\% & 5
& 0\% & 5
& 36\% & 1
& 28\% & 3
& \multicolumn{2}{>{\columncolor[HTML]{E7EEFE}}c}{%
    \begin{tabular}{@{}cc@{}}
        31\% & 2
    \end{tabular}
} \\
Move Playingcard Away
& 11\% & 4
& 0\%  & 6
& 0\%  & 6
& 3\%  & 5
& 22\% & 3
& 25\% & 2
& \multicolumn{2}{>{\columncolor[HTML]{E7EEFE}}c}{%
    \begin{tabular}{@{}cc@{}}
        32\% & 1
    \end{tabular}
} \\
Open Microwave
& 20\% & 5
& 0\% & 6
& 0\% & 6
& 22\% & 4
& 50\% & 1
& 39\% & 3
& \multicolumn{2}{>{\columncolor[HTML]{E7EEFE}}c}{%
    \begin{tabular}{@{}cc@{}}
        41\% & 2
    \end{tabular}
} \\
Pick Diverse Bottles
& 0\% & 4
& 0\% & 4
& 0\% & 4
& 1\% & 3
& 6\% & 2
& 6\% & 2
& \multicolumn{2}{>{\columncolor[HTML]{E7EEFE}}c}{%
    \begin{tabular}{@{}cc@{}}
     7\% & \ 1
    \end{tabular}
} \\
Turn Switch
& 15\% & 3
& 2\%  & 6
& 1\%  & 7
& 8\%  & 5
& 23\% & 1
& 11\% & 4
& \multicolumn{2}{>{\columncolor[HTML]{E7EEFE}}c}{%
    \begin{tabular}{@{}cc@{}}
        16\% & 2
    \end{tabular}
} \\
Place Object Stand
& 5\% & 3
& 0\%  & 4
& 0\%  & 4
& 0\%  & 4
& 11\% & 2
& 11\% & 2
& \multicolumn{2}{>{\columncolor[HTML]{E7EEFE}}c}{%
    \begin{tabular}{@{}cc@{}}
        13\% & 1
    \end{tabular}
} \\
Stack Blocks Two
& 2\% & 3
& 0\% & 5
& 0\% & 5
& 0\% & 5
& 1\% & 4
& 5\% & 2
& \multicolumn{2}{>{\columncolor[HTML]{E7EEFE}}c}{%
    \begin{tabular}{@{}cc@{}}
    11\% & 1
    \end{tabular}
} \\
\hline
Average
& 16\% & 4
& 1\%  & 6
& 1\%  & 6
& 8\% & 5
& 23\% & 3
& 24\% & 2
& \multicolumn{2}{>{\columncolor[HTML]{E7EEFE}}c}{%
    \begin{tabular}{@{}cc@{}}
        30\% & 1
    \end{tabular}
} \\
\bottomrule[1.2pt]
\end{tabular*}
\end{table*}

%% file: Supplementary/tb_real_world_generazation.tex
\begin{table*}[t]
  \centering
  \caption{Overview of \textit{Generalization Tasks} on Real-World.}
  \small
  \setlength{\tabcolsep}{4pt}
  \renewcommand{\arraystretch}{1.25}
  \begin{tabularx}{\textwidth}{l *{4}{Y}}
    \toprule[1.1pt]
    & \multicolumn{4}{c}{\textbf{Generalization Tasks}} \\
    \cmidrule(lr){2-5}
    \textbf{Task name}
      & place\_one\_fruit\_on\_plate
      & place\_plate\_on\_tablecloth
      & stand\_bottle\_upright
      & place\_cup\_on\_tablecloth \\
    \midrule
    \textbf{Description}
     & Grasp a piece of fruit and place it onto the plate. The set of graspable fruits includes bananas, apples, lemons, and related varieties.
      & Grasp a plate and place it on the tablecloth. Both the plates and the tablecloths come in multiple styles.
      & Grasp a water bottle and then orient it to an upright pose. The bottles exhibit diverse shapes and appearances.
      & Grasp a cup and place it on the tablecloth. The cups exhibit diverse shapes and appearances. \\
    \textbf{\# demonstrations}
      & 100 & 100 & 150 & 120 \\
    \textbf{\# rollouts}
      & 50 & 50 & 50 & 50 \\
    \textbf{Performance (\%)}
      & 88 & 90 & 74 & 88 \\
    \bottomrule[1.1pt]
  \end{tabularx}
  \label{tab:real-world-generalization}
\end{table*}

%% file: Supplementary/tb_real_world_long.tex
\begin{table*}[t]
  \centering
  \caption{Overview of \textit{Long-horizon Tasks} on Real-World.}
  \small
  \setlength{\tabcolsep}{4pt}
  \renewcommand{\arraystretch}{1.25}
  \begin{tabularx}{\textwidth}{l *{4}{Y}}
    \toprule[1.1pt]
    & \multicolumn{4}{c}{\textbf{Long-Horizon Tasks}} \\
    \cmidrule(lr){2-5}
    \textbf{Task name}
      & place\_all\_fruits\_on\_plate
      & place\_2\_fruits\_in\_2\_position
      & place\_2\_obj\_on\_plate
      & place\_blocks\_on\_tablecloth \\
    \midrule
    \textbf{Description}
      & Grasp three different fruits and place them onto the plate. For each rollout, a different combination of fruits is randomly sampled.
      & Grasp one fruit from the bowl and place it onto the tablecloth, then grasp another fruit from the plate and place it into the bowl.
      & Grasp a cup and an apple in sequence and place them into the plate.
      & Grasp three distinct blocks and place them on the tablecloth. For each rollout, a different combination of blocks is randomly sampled. \\
    \textbf{\# demonstrations}
      & 200 & 200 & 250 & 300 \\
    \textbf{\# rollouts}
      & 25 & 25 & 25 & 25 \\
    \textbf{Performance (\%)}
      & 64 & 60 & 68 & 64 \\
    \bottomrule[1.1pt]
  \end{tabularx}
  \label{tab:real-world-long-horizon}
\end{table*}

%% file: Supplementary/fig-case1.tex
\begin{figure*}[htbp]
    \centering
    \includegraphics[width=1\textwidth]{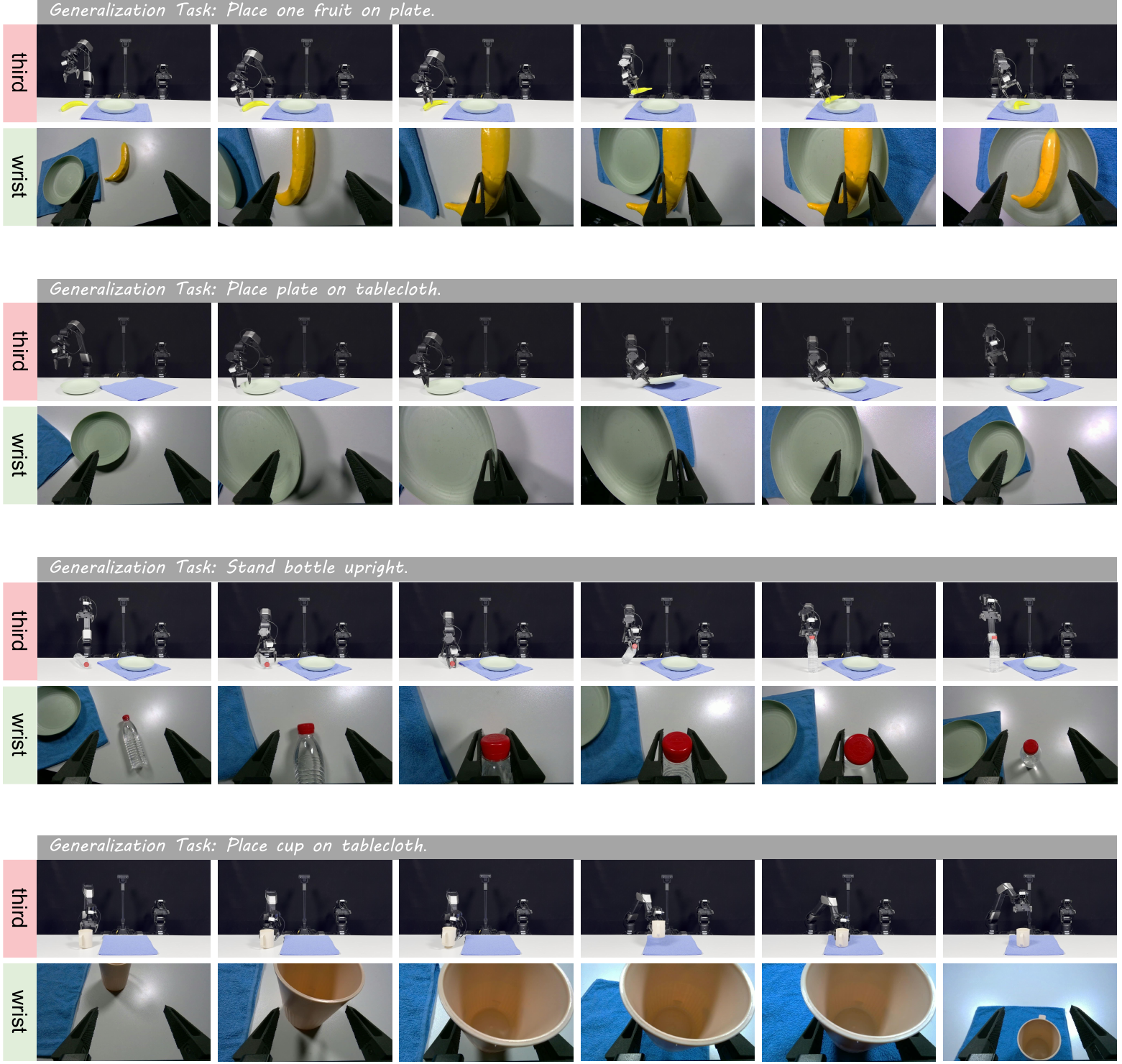}
    \caption{Qualitative results of OptimusVLA on Real-World. From top to bottom, we illustrate four \textit{Generalization Tasks}: \textit{Place one fruit on plate}, \textit{Place plate on tablecloth}, \textit{Stand bottle upright}, and \textit{Place cup on tablecloth}. Here, \texttt{third} denotes third-person view images, while \texttt{wrist} denotes images captured from the robot’s wrist-mounted camera.}
    \label{fig:appendix-case1}
\end{figure*}

%% file: Supplementary/fig-case2.tex
\begin{figure*}[htbp]
    \centering
    \includegraphics[width=1\textwidth]{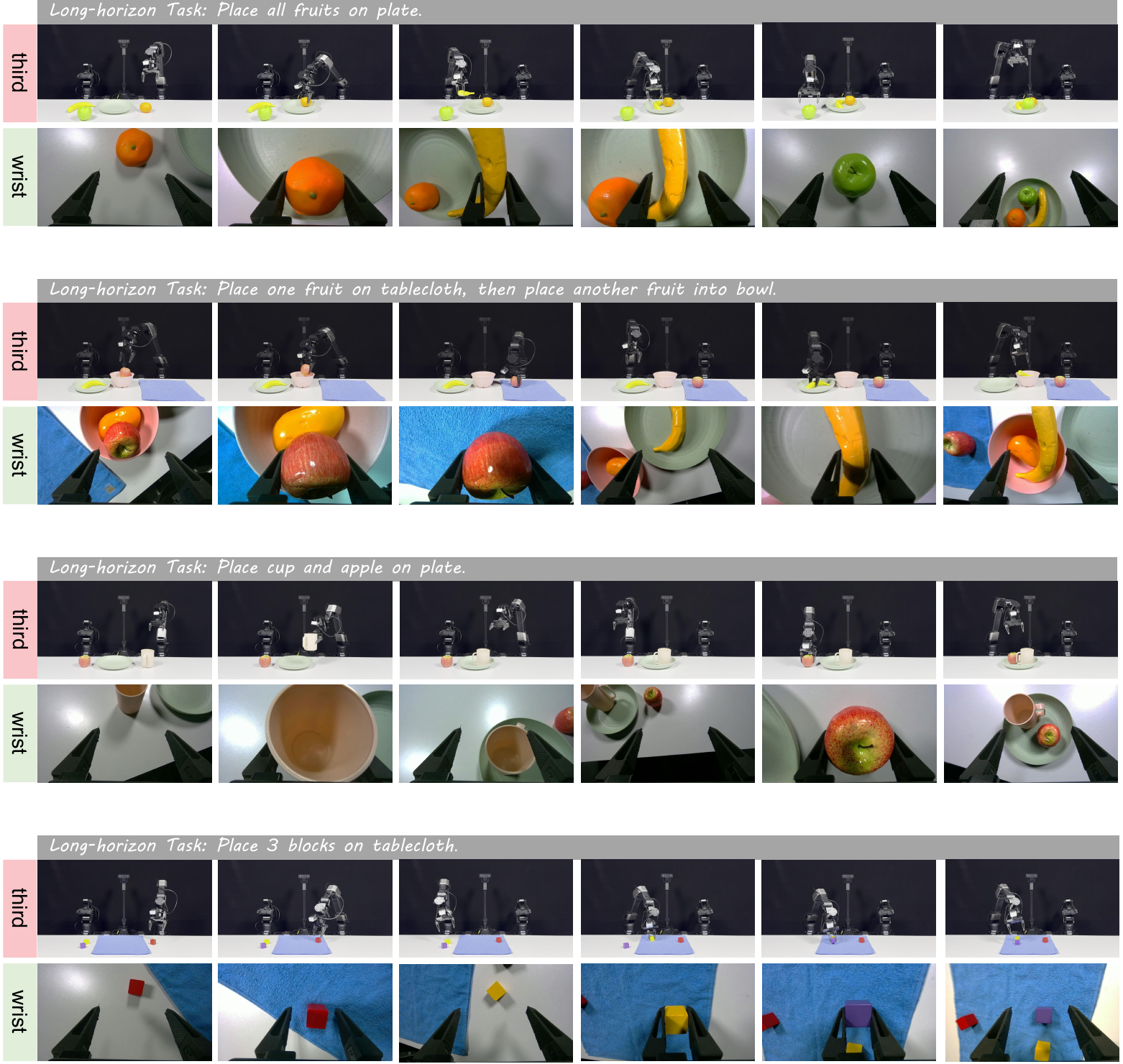}
    \caption{Qualitative results of OptimusVLA on Real-World. From top to bottom, we illustrate four \textit{Long-horizon Tasks}: \textit{Place all fruits on plate}, \textit{Place one fruit on tablecloth, then place another fruit into bowl}, \textit{Place cup and apple on plate}, and \textit{Place 3 blocks on tablecloth}. Here, \texttt{third} denotes third-person view images, while \texttt{wrist} denotes images captured from the robot’s wrist-mounted camera.}
    \label{fig:appendix-case2}
\end{figure*}

%% file: main.bbl
\begin{thebibliography}{61}
\providecommand{\natexlab}[1]{#1}
\providecommand{\url}[1]{\texttt{#1}}
\expandafter\ifx\csname urlstyle\endcsname\relax
  \providecommand{\doi}[1]{doi: #1}\else
  \providecommand{\doi}{doi: \begingroup \urlstyle{rm}\Url}\fi

\bibitem[Black et~al.(2024)Black, Brown, Driess, Esmail, Equi, Finn, Fusai, Groom, Hausman, Ichter, et~al.]{black2024pi0}
Kevin Black, Noah Brown, Danny Driess, Adnan Esmail, Michael Equi, Chelsea Finn, Niccolo Fusai, Lachy Groom, Karol Hausman, Brian Ichter, et~al.
\newblock $pi\_0$: A vision-language-action flow model for general robot control.
\newblock \emph{arXiv preprint arXiv:2410.24164}, 2024.

\bibitem[Bu et~al.(2024)Bu, Li, Chen, Cai, Zeng, Cui, Yao, and Qiao]{bu2024towards}
Qingwen Bu, Hongyang Li, Li Chen, Jisong Cai, Jia Zeng, Heming Cui, Maoqing Yao, and Yu Qiao.
\newblock Towards synergistic, generalized, and efficient dual-system for robotic manipulation.
\newblock \emph{arXiv preprint arXiv:2410.08001}, 2024.

\bibitem[Bu et~al.(2025)Bu, Yang, Cai, Gao, Ren, Yao, Luo, and Li]{bu2025univla}
Qingwen Bu, Yanting Yang, Jisong Cai, Shenyuan Gao, Guanghui Ren, Maoqing Yao, Ping Luo, and Hongyang Li.
\newblock Univla: Learning to act anywhere with task-centric latent actions.
\newblock \emph{arXiv preprint arXiv:2505.06111}, 2025.

\bibitem[Chen et~al.(2024)Chen, Shen, Shao, Deng, and Nie]{chen2024lion}
Gongwei Chen, Leyang Shen, Rui Shao, Xiang Deng, and Liqiang Nie.
\newblock Lion: Empowering multimodal large language model with dual-level visual knowledge.
\newblock In \emph{Proceedings of the IEEE/CVF Conference on Computer Vision and Pattern Recognition}, pages 26540--26550, 2024.

\bibitem[Chen et~al.(2025)Chen, Chen, Chen, Cai, Liu, Li, Liang, Lin, Ge, Gu, et~al.]{chen2025robotwin2}
Tianxing Chen, Zanxin Chen, Baijun Chen, Zijian Cai, Yibin Liu, Zixuan Li, Qiwei Liang, Xianliang Lin, Yiheng Ge, Zhenyu Gu, et~al.
\newblock Robotwin 2.0: A scalable data generator and benchmark with strong domain randomization for robust bimanual robotic manipulation.
\newblock \emph{arXiv preprint arXiv:2506.18088}, 2025.

\bibitem[Chi et~al.(2025)Chi, Xu, Feng, Cousineau, Du, Burchfiel, Tedrake, and Song]{chi2025diffusionpolicy}
Cheng Chi, Zhenjia Xu, Siyuan Feng, Eric Cousineau, Yilun Du, Benjamin Burchfiel, Russ Tedrake, and Shuran Song.
\newblock Diffusion policy: Visuomotor policy learning via action diffusion.
\newblock \emph{The International Journal of Robotics Research}, 44\penalty0 (10-11):\penalty0 1684--1704, 2025.

\bibitem[Deng et~al.(2025)Deng, Yan, Wei, Ma, Yang, Chen, Zhang, Yang, Zhang, Zhang, et~al.]{deng2025graspvla}
Shengliang Deng, Mi Yan, Songlin Wei, Haixin Ma, Yuxin Yang, Jiayi Chen, Zhiqi Zhang, Taoyu Yang, Xuheng Zhang, Wenhao Zhang, et~al.
\newblock Graspvla: a grasping foundation model pre-trained on billion-scale synthetic action data.
\newblock \emph{arXiv preprint arXiv:2505.03233}, 2025.

\bibitem[Fan et~al.(2025)Fan, Jia, Sun, Wang, Wei, Gong, Zhao, Tomizuka, Yang, Yan, et~al.]{fan2025interleave-vla}
Cunxin Fan, Xiaosong Jia, Yihang Sun, Yixiao Wang, Jianglan Wei, Ziyang Gong, Xiangyu Zhao, Masayoshi Tomizuka, Xue Yang, Junchi Yan, et~al.
\newblock Interleave-vla: Enhancing robot manipulation with interleaved image-text instructions.
\newblock \emph{arXiv preprint arXiv:2505.02152}, 2025.

\bibitem[Gu and Dao(2024)]{gu2024mamba}
Albert Gu and Tri Dao.
\newblock Mamba: Linear-time sequence modeling with selective state spaces.
\newblock In \emph{First conference on language modeling}, 2024.

\bibitem[Hu et~al.(2024)Hu, Guo, Wang, Chen, Wang, Zhang, Sreenath, Lu, and Chen]{hu2024video}
Yucheng Hu, Yanjiang Guo, Pengchao Wang, Xiaoyu Chen, Yen-Jen Wang, Jianke Zhang, Koushil Sreenath, Chaochao Lu, and Jianyu Chen.
\newblock Video prediction policy: A generalist robot policy with predictive visual representations.
\newblock \emph{arXiv preprint arXiv:2412.14803}, 2024.

\bibitem[Huang et~al.(2023)Huang, Yong, Ma, Linghu, Li, Wang, Li, Zhu, Jia, and Huang]{huang2023leo}
Jiangyong Huang, Silong Yong, Xiaojian Ma, Xiongkun Linghu, Puhao Li, Yan Wang, Qing Li, Song-Chun Zhu, Baoxiong Jia, and Siyuan Huang.
\newblock An embodied generalist agent in 3d world.
\newblock \emph{arXiv preprint arXiv:2311.12871}, 2023.

\bibitem[Intelligence et~al.(2025)Intelligence, Black, Brown, Darpinian, Dhabalia, Driess, Esmail, Equi, Finn, Fusai, et~al.]{intelligence2025pi05}
Physical Intelligence, Kevin Black, Noah Brown, James Darpinian, Karan Dhabalia, Danny Driess, Adnan Esmail, Michael Equi, Chelsea Finn, Niccolo Fusai, et~al.
\newblock pi05: a vision-language-action model with open-world generalization.
\newblock \emph{arXiv preprint arXiv:2504.16054}, 2025.

\bibitem[Jiang et~al.(2022)Jiang, Gupta, Zhang, Wang, Dou, Chen, Fei-Fei, Anandkumar, Zhu, and Fan]{jiang2022vima}
Yunfan Jiang, Agrim Gupta, Zichen Zhang, Guanzhi Wang, Yongqiang Dou, Yanjun Chen, Li Fei-Fei, Anima Anandkumar, Yuke Zhu, and Linxi Fan.
\newblock Vima: General robot manipulation with multimodal prompts.
\newblock In \emph{NeurIPS 2022 Foundation Models for Decision Making Workshop}, 2022.

\bibitem[Kim et~al.(2024)Kim, Pertsch, Karamcheti, Xiao, Balakrishna, Nair, Rafailov, Foster, Lam, Sanketi, et~al.]{kim2024openvla}
Moo~Jin Kim, Karl Pertsch, Siddharth Karamcheti, Ted Xiao, Ashwin Balakrishna, Suraj Nair, Rafael Rafailov, Ethan Foster, Grace Lam, Pannag Sanketi, et~al.
\newblock Openvla: An open-source vision-language-action model.
\newblock \emph{arXiv preprint arXiv:2406.09246}, 2024.

\bibitem[Kim et~al.(2025)Kim, Finn, and Liang]{kim2025openvla-oft}
Moo~Jin Kim, Chelsea Finn, and Percy Liang.
\newblock Fine-tuning vision-language-action models: Optimizing speed and success.
\newblock \emph{arXiv preprint arXiv:2502.19645}, 2025.

\bibitem[Li et~al.(2025{\natexlab{a}})Li, Lv, Shao, Deng, Li, Hao, and Nie]{li2025star}
Hao Li, Qi Lv, Rui Shao, Xiang Deng, Yinchuan Li, Jianye Hao, and Liqiang Nie.
\newblock Star: Learning diverse robot skill abstractions through rotation-augmented vector quantization.
\newblock In \emph{International Conference on Machine Learning}, 2025{\natexlab{a}}.

\bibitem[Li et~al.(2024{\natexlab{a}})Li, Liang, Wang, Luo, Chen, Liao, Wei, Deng, Xu, Zhang, et~al.]{li2024cogact}
Qixiu Li, Yaobo Liang, Zeyu Wang, Lin Luo, Xi Chen, Mozheng Liao, Fangyun Wei, Yu Deng, Sicheng Xu, Yizhong Zhang, et~al.
\newblock Cogact: A foundational vision-language-action model for synergizing cognition and action in robotic manipulation.
\newblock \emph{arXiv preprint arXiv:2411.19650}, 2024{\natexlab{a}}.

\bibitem[Li et~al.(2025{\natexlab{b}})Li, Zhang, Shao, He, and Nie]{li2025cogvla}
Wei Li, Renshan Zhang, Rui Shao, Jie He, and Liqiang Nie.
\newblock Cogvla: Cognition-aligned vision-language-action model via instruction-driven routing \& sparsification.
\newblock \emph{arXiv preprint arXiv:2508.21046}, 2025{\natexlab{b}}.

\bibitem[Li et~al.(2026)Li, Zhang, Shao, Fang, Zhou, Tian, and Nie]{li2025semanticvla}
Wei Li, Renshan Zhang, Rui Shao, Zhijian Fang, Kaiwen Zhou, Zhuotao Tian, and Liqiang Nie.
\newblock Semanticvla: Semantic-aligned sparsification and enhancement for efficient robotic manipulation.
\newblock In \emph{Proceedings of the AAAI Conference on Artificial Intelligence}, 2026.

\bibitem[Li et~al.(2023)Li, Liu, Zhang, Yu, Xu, Wu, Cheang, Jing, Zhang, Liu, et~al.]{li2023roboflamingo}
Xinghang Li, Minghuan Liu, Hanbo Zhang, Cunjun Yu, Jie Xu, Hongtao Wu, Chilam Cheang, Ya Jing, Weinan Zhang, Huaping Liu, et~al.
\newblock Vision-language foundation models as effective robot imitators.
\newblock \emph{arXiv preprint arXiv:2311.01378}, 2023.

\bibitem[Li et~al.(2022)Li, Tang, Zhao, and Zhu]{li2022emocaps}
Zaijing Li, Fengxiao Tang, Ming Zhao, and Yusen Zhu.
\newblock {E}mo{C}aps: Emotion capsule based model for conversational emotion recognition.
\newblock In \emph{Findings of the Association for Computational Linguistics: ACL 2022}, pages 1610--1618. Association for Computational Linguistics, 2022.

\bibitem[Li et~al.(2024{\natexlab{b}})Li, Xie, Shao, Chen, Jiang, and Nie]{li2024optimus}
Zaijing Li, Yuquan Xie, Rui Shao, Gongwei Chen, Dongmei Jiang, and Liqiang Nie.
\newblock Optimus-1: Hybrid multimodal memory empowered agents excel in long-horizon tasks.
\newblock \emph{arXiv preprint arXiv:2408.03615}, 2024{\natexlab{b}}.

\bibitem[Li et~al.(2025{\natexlab{c}})Li, Xie, Shao, Chen, Guan, Jiang, Wang, and Nie]{li2026optimus3}
Zaijing Li, Yuquan Xie, Rui Shao, Gongwei Chen, Weili Guan, Dongmei Jiang, Yaowei Wang, and Liqiang Nie.
\newblock Optimus-3: Dual-router aligned mixture-of-experts agent with dual-granularity reasoning-aware policy optimization.
\newblock \emph{arXiv preprint arXiv:2506.10357}, 2025{\natexlab{c}}.

\bibitem[Li et~al.(2025{\natexlab{d}})Li, Xie, Shao, Chen, Jiang, and Nie]{li2025optimus2}
Zaijing Li, Yuquan Xie, Rui Shao, Gongwei Chen, Dongmei Jiang, and Liqiang Nie.
\newblock Optimus-2: Multimodal minecraft agent with goal-observation-action conditioned policy.
\newblock In \emph{Proceedings of the IEEE/CVF Conference on Computer Vision and Pattern Recognition (CVPR)}, pages 9039--9049, 2025{\natexlab{d}}.

\bibitem[Lipman et~al.(2022)Lipman, Chen, Ben-Hamu, Nickel, and Le]{lipman2022flow}
Yaron Lipman, Ricky~TQ Chen, Heli Ben-Hamu, Maximilian Nickel, and Matt Le.
\newblock Flow matching for generative modeling.
\newblock \emph{arXiv preprint arXiv:2210.02747}, 2022.

\bibitem[Liu et~al.(2023)Liu, Zhu, Gao, Feng, Liu, Zhu, and Stone]{liu2023libero}
Bo Liu, Yifeng Zhu, Chongkai Gao, Yihao Feng, Qiang Liu, Yuke Zhu, and Peter Stone.
\newblock Libero: Benchmarking knowledge transfer for lifelong robot learning.
\newblock \emph{Advances in Neural Information Processing Systems}, 36:\penalty0 44776--44791, 2023.

\bibitem[Liu et~al.(2024{\natexlab{a}})Liu, Li, Wu, and Lee]{liu2024visual}
Haotian Liu, Chunyuan Li, Qingyang Wu, and Yong~Jae Lee.
\newblock Visual instruction tuning.
\newblock \emph{Advances in neural information processing systems}, 36, 2024{\natexlab{a}}.

\bibitem[Liu et~al.(2025{\natexlab{a}})Liu, Li, Li, Liu, Wang, Liu, Kang, Ma, Kong, and Zhang]{liu2025robovlm}
Huaping Liu, Xinghang Li, Peiyan Li, Minghuan Liu, Dong Wang, Jirong Liu, Bingyi Kang, Xiao Ma, Tao Kong, and Hanbo Zhang.
\newblock Towards generalist robot policies: What matters in building vision-language-action models.
\newblock 2025{\natexlab{a}}.

\bibitem[Liu et~al.(2025{\natexlab{b}})Liu, Li, Li, Liu, Wang, Liu, Kang, Ma, Kong, and Zhang]{liu2025robovlms}
Huaping Liu, Xinghang Li, Peiyan Li, Minghuan Liu, Dong Wang, Jirong Liu, Bingyi Kang, Xiao Ma, Tao Kong, and Hanbo Zhang.
\newblock Towards generalist robot policies: What matters in building vision-language-action models.
\newblock 2025{\natexlab{b}}.

\bibitem[Liu et~al.(2025{\natexlab{c}})Liu, Chen, An, Liu, Zhang, Gu, Li, Guo, Chen, Liu, et~al.]{liu2025hybridvla}
Jiaming Liu, Hao Chen, Pengju An, Zhuoyang Liu, Renrui Zhang, Chenyang Gu, Xiaoqi Li, Ziyu Guo, Sixiang Chen, Mengzhen Liu, et~al.
\newblock Hybridvla: Collaborative diffusion and autoregression in a unified vision-language-action model.
\newblock \emph{arXiv preprint arXiv:2503.10631}, 2025{\natexlab{c}}.

\bibitem[Liu et~al.(2024{\natexlab{b}})Liu, Wu, Li, Tan, Chen, Wang, Xu, Su, and Zhu]{liu2024rdt}
Songming Liu, Lingxuan Wu, Bangguo Li, Hengkai Tan, Huayu Chen, Zhengyi Wang, Ke Xu, Hang Su, and Jun Zhu.
\newblock Rdt-1b: a diffusion foundation model for bimanual manipulation.
\newblock \emph{arXiv preprint arXiv:2410.07864}, 2024{\natexlab{b}}.

\bibitem[Liu et~al.(2022)Liu, Gong, and Liu]{liu2022flow}
Xingchao Liu, Chengyue Gong, and Qiang Liu.
\newblock Flow straight and fast: Learning to generate and transfer data with rectified flow.
\newblock \emph{arXiv preprint arXiv:2209.03003}, 2022.

\bibitem[Ma et~al.(2024)Ma, Patidar, Haughton, and James]{ma2024hierarchical}
Xiao Ma, Sumit Patidar, Iain Haughton, and Stephen James.
\newblock Hierarchical diffusion policy for kinematics-aware multi-task robotic manipulation.
\newblock In \emph{Proceedings of the IEEE/CVF Conference on Computer Vision and Pattern Recognition}, pages 18081--18090, 2024.

\bibitem[Mees et~al.(2022)Mees, Hermann, Rosete-Beas, and Burgard]{mees2022calvin}
Oier Mees, Lukas Hermann, Erick Rosete-Beas, and Wolfram Burgard.
\newblock Calvin: A benchmark for language-conditioned policy learning for long-horizon robot manipulation tasks.
\newblock \emph{IEEE Robotics and Automation Letters}, 7\penalty0 (3):\penalty0 7327--7334, 2022.

\bibitem[Ortenzi et~al.(2018)Ortenzi, Marturi, Mistry, Kuo, and Stolkin]{Ortenzi2018kinematics}
Valerio Ortenzi, Naresh Marturi, Michael Mistry, Jeffrey Kuo, and Rustam Stolkin.
\newblock Vision-based framework to estimate robot configuration and kinematic constraints.
\newblock \emph{IEEE/ASME Transactions on Mechatronics}, 23\penalty0 (5):\penalty0 2402--2412, 2018.

\bibitem[Pertsch et~al.(2025)Pertsch, Stachowicz, Ichter, Driess, Nair, Vuong, Mees, Finn, and Levine]{pertsch2025pi0fast}
Karl Pertsch, Kyle Stachowicz, Brian Ichter, Danny Driess, Suraj Nair, Quan Vuong, Oier Mees, Chelsea Finn, and Sergey Levine.
\newblock Fast: Efficient action tokenization for vision-language-action models.
\newblock \emph{arXiv preprint arXiv:2501.09747}, 2025.

\bibitem[Qu et~al.(2025)Qu, Song, Chen, Yao, Ye, Ding, Wang, Gu, Zhao, Wang, et~al.]{qu2025spatialvla}
Delin Qu, Haoming Song, Qizhi Chen, Yuanqi Yao, Xinyi Ye, Yan Ding, Zhigang Wang, JiaYuan Gu, Bin Zhao, Dong Wang, et~al.
\newblock Spatialvla: Exploring spatial representations for visual-language-action model.
\newblock \emph{arXiv preprint arXiv:2501.15830}, 2025.

\bibitem[Shi et~al.(2025)Shi, Xie, Liu, Sun, Liu, Wang, Zhou, Fan, Zhang, and Huang]{shi2025memoryvla}
Hao Shi, Bin Xie, Yingfei Liu, Lin Sun, Fengrong Liu, Tiancai Wang, Erjin Zhou, Haoqiang Fan, Xiangyu Zhang, and Gao Huang.
\newblock Memoryvla: Perceptual-cognitive memory in vision-language-action models for robotic manipulation.
\newblock \emph{arXiv preprint arXiv:2508.19236}, 2025.

\bibitem[Song et~al.(2026)Song, Deng, Wei, Jiang, Nie, and Guan]{song2026energyaction}
Mingchen Song, Xiang Deng, Jie Wei, Dongmei Jiang, Liqiang Nie, and Weili Guan.
\newblock Energyaction: Unimanual to bimanual composition with energy-based models.
\newblock \emph{arXiv preprint arXiv:2603.20236}, 2026.

\bibitem[Song et~al.(2025{\natexlab{a}})Song, Chen, Ding, Zhao, Zhao, Zhong, Ge, Ma, and Li]{song2025pdvla}
Wenxuan Song, Jiayi Chen, Pengxiang Ding, Han Zhao, Wei Zhao, Zhide Zhong, Zongyuan Ge, Jun Ma, and Haoang Li.
\newblock Accelerating vision-language-action model integrated with action chunking via parallel decoding.
\newblock \emph{arXiv preprint arXiv:2503.02310}, 2025{\natexlab{a}}.

\bibitem[Song et~al.(2025{\natexlab{b}})Song, Zhou, Zhao, Chen, Ding, Yan, Huang, Tang, Wang, and Li]{song2025reconvla}
Wenxuan Song, Ziyang Zhou, Han Zhao, Jiayi Chen, Pengxiang Ding, Haodong Yan, Yuxin Huang, Feilong Tang, Donglin Wang, and Haoang Li.
\newblock Reconvla: Reconstructive vision-language-action model as effective robot perceiver.
\newblock \emph{arXiv preprint arXiv:2508.10333}, 2025{\natexlab{b}}.

\bibitem[Team et~al.(2024)Team, Ghosh, Walke, Pertsch, Black, Mees, Dasari, Hejna, Kreiman, Xu, et~al.]{team2024octo}
Octo~Model Team, Dibya Ghosh, Homer Walke, Karl Pertsch, Kevin Black, Oier Mees, Sudeep Dasari, Joey Hejna, Tobias Kreiman, Charles Xu, et~al.
\newblock Octo: An open-source generalist robot policy.
\newblock \emph{arXiv preprint arXiv:2405.12213}, 2024.

\bibitem[Tian et~al.(2024)Tian, Yang, Zeng, Wang, Lin, Dong, and Pang]{tian2024predictive}
Yang Tian, Sizhe Yang, Jia Zeng, Ping Wang, Dahua Lin, Hao Dong, and Jiangmiao Pang.
\newblock Predictive inverse dynamics models are scalable learners for robotic manipulation.
\newblock \emph{arXiv preprint arXiv:2412.15109}, 2024.

\bibitem[Vaswani et~al.(2017)Vaswani, Shazeer, Parmar, Uszkoreit, Jones, Gomez, Kaiser, and Polosukhin]{attention2017}
Ashish Vaswani, Noam Shazeer, Niki Parmar, Jakob Uszkoreit, Llion Jones, Aidan~N. Gomez, \L{}ukasz Kaiser, and Illia Polosukhin.
\newblock Attention is all you need.
\newblock page 6000–6010, 2017.

\bibitem[Wang et~al.(2025{\natexlab{a}})Wang, Xiong, Wang, and Chen]{wang2025bitvla}
Hongyu Wang, Chuyan Xiong, Ruiping Wang, and Xilin Chen.
\newblock Bitvla: 1-bit vision-language-action models for robotics manipulation.
\newblock \emph{arXiv preprint arXiv:2506.07530}, 2025{\natexlab{a}}.

\bibitem[Wang et~al.(2025{\natexlab{b}})Wang, Ding, Li, Cui, Ge, Tong, Song, Zhao, Zhao, Hou, et~al.]{wang2025vla-adapter}
Yihao Wang, Pengxiang Ding, Lingxiao Li, Can Cui, Zirui Ge, Xinyang Tong, Wenxuan Song, Han Zhao, Wei Zhao, Pengxu Hou, et~al.
\newblock Vla-adapter: An effective paradigm for tiny-scale vision-language-action model.
\newblock \emph{arXiv preprint arXiv:2509.09372}, 2025{\natexlab{b}}.

\bibitem[Wen et~al.(2025)Wen, Zhu, Li, Zhu, Tang, Wu, Xu, Liu, Cheng, Shen, et~al.]{wen2025tinyvla}
Junjie Wen, Yichen Zhu, Jinming Li, Minjie Zhu, Zhibin Tang, Kun Wu, Zhiyuan Xu, Ning Liu, Ran Cheng, Chaomin Shen, et~al.
\newblock Tinyvla: Towards fast, data-efficient vision-language-action models for robotic manipulation.
\newblock \emph{IEEE Robotics and Automation Letters}, 2025.

\bibitem[Xiang et~al.(2026)Xiang, Zhang, Zhang, Wang, Hou, and Nie]{xiang2026tina}
Qianlong Xiang, Miao Zhang, Haoyu Zhang, Kun Wang, Junhui Hou, and Liqiang Nie.
\newblock Tina: Text-free inversion attack for unlearned text-to-image diffusion models.
\newblock \emph{arXiv preprint arXiv:2603.17828}, 2026.

\bibitem[Ze et~al.(2024)Ze, Zhang, Zhang, Hu, Wang, and Xu]{ze2024dp3}
Yanjie Ze, Gu Zhang, Kangning Zhang, Chenyuan Hu, Muhan Wang, and Huazhe Xu.
\newblock 3d diffusion policy: Generalizable visuomotor policy learning via simple 3d representations.
\newblock \emph{arXiv preprint arXiv:2403.03954}, 2024.

\bibitem[Zhang et~al.(2023)Zhang, Liu, Li, Yan, Gao, Chang, and Nie]{zhang2023attribute}
Haoyu Zhang, Meng Liu, Yuhong Li, Ming Yan, Zan Gao, Xiaojun Chang, and Liqiang Nie.
\newblock Attribute-guided collaborative learning for partial person re-identification.
\newblock \emph{IEEE Transactions on Pattern Analysis and Machine Intelligence}, 45\penalty0 (12):\penalty0 14144--14160, 2023.

\bibitem[Zhang et~al.(2024)Zhang, Liu, Liu, Song, Wang, and Nie]{zhang2024multi}
Haoyu Zhang, Meng Liu, Zixin Liu, Xuemeng Song, Yaowei Wang, and Liqiang Nie.
\newblock Multi-factor adaptive vision selection for egocentric video question answering.
\newblock In \emph{Forty-first International Conference on Machine Learning}, pages 59310--59328, 2024.

\bibitem[Zhang et~al.(2025{\natexlab{a}})Zhang, Liu, Li, Wen, Guan, Wang, and Nie]{zhang2025spatial}
Haoyu Zhang, Meng Liu, Zaijing Li, Haokun Wen, Weili Guan, Yaowei Wang, and Liqiang Nie.
\newblock Spatial understanding from videos: Structured prompts meet simulation data.
\newblock In \emph{Advances in Neural Information Processing Systems}, pages 1--16, 2025{\natexlab{a}}.

\bibitem[Zhang et~al.(2025{\natexlab{b}})Zhang, Guo, Hu, Chen, Zhu, and Chen]{zhang2025up}
Jianke Zhang, Yanjiang Guo, Yucheng Hu, Xiaoyu Chen, Xiang Zhu, and Jianyu Chen.
\newblock Up-vla: A unified understanding and prediction model for embodied agent.
\newblock \emph{arXiv preprint arXiv:2501.18867}, 2025{\natexlab{b}}.

\bibitem[Zhang et~al.(2025{\natexlab{c}})Zhang, Liu, Fan, Liu, Zeng, and Liu]{zhang2025flowpolicy}
Qinglun Zhang, Zhen Liu, Haoqiang Fan, Guanghui Liu, Bing Zeng, and Shuaicheng Liu.
\newblock Flowpolicy: Enabling fast and robust 3d flow-based policy via consistency flow matching for robot manipulation.
\newblock In \emph{Proceedings of the AAAI Conference on Artificial Intelligence}, pages 14754--14762, 2025{\natexlab{c}}.

\bibitem[Zhang et~al.(2025{\natexlab{d}})Zhang, Shao, Chen, Zhang, Zhou, Guan, and Nie]{zhang2025falcon}
Renshan Zhang, Rui Shao, Gongwei Chen, Miao Zhang, Kaiwen Zhou, Weili Guan, and Liqiang Nie.
\newblock Falcon: Resolving visual redundancy and fragmentation in high-resolution multimodal large language models via visual registers.
\newblock In \emph{Proceedings of the IEEE/CVF International Conference on Computer Vision}, pages 23530--23540, 2025{\natexlab{d}}.

\bibitem[Zhang et~al.(2025{\natexlab{e}})Zhang, Liu, Qi, Wang, Yu, Zhang, Dong, He, Lu, Wang, et~al.]{zhang2025dreamvla}
Wenyao Zhang, Hongsi Liu, Zekun Qi, Yunnan Wang, Xinqiang Yu, Jiazhao Zhang, Runpei Dong, Jiawei He, Fan Lu, He Wang, et~al.
\newblock Dreamvla: a vision-language-action model dreamed with comprehensive world knowledge.
\newblock \emph{arXiv preprint arXiv:2507.04447}, 2025{\natexlab{e}}.

\bibitem[Zhao et~al.(2023)Zhao, Kumar, Levine, and Finn]{zhao2023act}
Tony~Z Zhao, Vikash Kumar, Sergey Levine, and Chelsea Finn.
\newblock Learning fine-grained bimanual manipulation with low-cost hardware.
\newblock \emph{arXiv preprint arXiv:2304.13705}, 2023.

\bibitem[Zheng et~al.(2024)Zheng, Liang, Huang, Gao, Daum{\'e}~III, Kolobov, Huang, and Yang]{zheng2024tracevla}
Ruijie Zheng, Yongyuan Liang, Shuaiyi Huang, Jianfeng Gao, Hal Daum{\'e}~III, Andrey Kolobov, Furong Huang, and Jianwei Yang.
\newblock Tracevla: Visual trace prompting enhances spatial-temporal awareness for generalist robotic policies.
\newblock \emph{arXiv preprint arXiv:2412.10345}, 2024.

\bibitem[Zhou et~al.(2025)Zhou, Chen, Xie, Li, Zhou, Wang, Yang, Tian, and Shao]{zhou2025hiconagent}
Xurui Zhou, Gongwei Chen, Yuquan Xie, Zaijing Li, Kaiwen Zhou, Shuai Wang, Shuo Yang, Zhuotao Tian, and Rui Shao.
\newblock Hiconagent: History context-aware policy optimization for gui agents.
\newblock \emph{arXiv preprint arXiv:2512.01763}, 2025.

\bibitem[Zhu et~al.(2026)Zhu, Shao, Liu, He, Liu, Wang, and Yu]{zhu2026H-GAR}
Yijie Zhu, Rui Shao, Ziyang Liu, Jie He, Jizhihui Liu, Jiuru Wang, and Zitong Yu.
\newblock H-gar: A hierarchical interaction framework via goal-driven observation-action refinement for robotic manipulation.
\newblock In \emph{Proceedings of the AAAI Conference on Artificial Intelligence}, 2026.

\bibitem[Zitkovich et~al.(2023)Zitkovich, Yu, Xu, Xu, Xiao, Xia, Wu, Wohlhart, Welker, Wahid, et~al.]{zitkovich2023rt2}
Brianna Zitkovich, Tianhe Yu, Sichun Xu, Peng Xu, Ted Xiao, Fei Xia, Jialin Wu, Paul Wohlhart, Stefan Welker, Ayzaan Wahid, et~al.
\newblock Rt-2: Vision-language-action models transfer web knowledge to robotic control.
\newblock In \emph{Conference on Robot Learning}, pages 2165--2183. PMLR, 2023.

\end{thebibliography}
